\documentclass{LMCS}

\def\doi{8(3:30)2012}
\lmcsheading%
{\doi}
{1--39}
{}
{}
{Mar.~\phantom08, 2011}
{Sep.~30, 2012}
{}

\usepackage{xspace}
\usepackage{tikz}
\usepackage{amsmath}
\usepackage{url}
\usepackage{enumerate,hyperref}

\newcommand{\ursa}{{\sc ursa}\xspace}
\newcommand{\sat}{{\sc sat}\xspace}
\newcommand{\dpll}{{\sc dpll}\xspace}
\newcommand{\argosat}{{\sc ArgoSAT}\xspace}
\newcommand{\csp}{{\sc csp}\xspace}
\newcommand{\smt}{{\sc smt}\xspace}
\newcommand{\np}{{\sc np}\xspace}
\newcommand{\cnf}{{\sc cnf}\xspace}
\newcommand{\dimacs}{{\sc dimacs}\xspace}

\newcommand{\des}{{\sc des}\xspace}

\newcommand{\NOT}{\neg}
\newcommand{\AND}{\wedge}
\newcommand{\OR}{\vee}
\newcommand{\XOR}{\oplus}
\newcommand{\IMPL}{\Rightarrow}
\newcommand{\EQ}{\Leftrightarrow}

\begin{document}

\title[URSA: A System for Uniform Reduction to SAT]{URSA: A System for Uniform Reduction to SAT}
\author[P.~Jani\v{c}i\'c]{Predrag Jani\v{c}i\'c}
\address{Faculty of Mathematics, Studentski trg 16, 11000 Belgrade, Serbia}
\email{janicic@matf.bg.ac.rs}

\begin{abstract}
There are a huge number of problems, from various areas, being solved by
reducing them to \sat. However, for many applications, translation into
\sat is performed by specialized, problem-specific tools. In this paper
we describe a new system for uniform solving of a wide class of problems
by reducing them to \sat. The system uses a new specification language
\ursa that combines imperative and declarative programming paradigms.
The reduction to \sat is defined precisely by the semantics of the specification
language. The domain of the approach is wide (e.g., many {\sc np}-complete
problems can be simply specified and then solved by the system) and there
are problems easily solvable by the proposed system, while they can be
hardly solved by using other programming languages or constraint programming
systems. So, the system can be seen not only as a tool for solving problems
by reducing them to \sat, but also as a general-purpose constraint solving
system (for finite domains). In this paper, we also describe an open-source
implementation of the described approach. The performed experiments suggest
that the system is competitive to state-of-the-art related modelling systems.
\end{abstract}


\maketitle

% ***************************************************************************
% ***************************************************************************
\section{Introduction}
\label{sec:introduction}
% ***************************************************************************
% ***************************************************************************

Following spectacular advances made over the last years, the \sat solving
technology has many successful applications nowadays --- there is a wide range
of problems being solved by reducing them to \sat. Very often solving based on
reduction to \sat is more efficient than using a problem-specific solution.
Therefore, \sat solvers are already considered to be a {\em Swiss army knife}
for solving many hard \csp and \np-complete problems and in many areas including
software and hardware verification, model checking, termination analysis,
planning, scheduling, cryptanalysis, electronic design automation, etc.
\cite{BordeauxHZ06,HandbookOfSAT2009,ClarkeBRZ01,termination-SAT,scheduling-SAT,Massacci00,MironovZ06}.
Typically, translations into \sat are performed by specialized,
problem-specific tools. However, using a general-purpose system capable of
reducing a wide range of problems to \sat can simplify this task,
make it less error prone, and make this approach more easily accessible
and more widely accepted.

There are already a number of approaches for solving combinatorial and
related problems by general-purpose systems that reduce problems to underlying
theories and domains (instead of developing special purpose
algorithms and implementations). A common motivation is that it is much easier
to develop a problem specification for a general system than a new,
special-purpose solver. The general problem solving systems include
libraries for general purpose programming languages, but also modelling and
programming languages built on top of specific solvers
\cite{ilog-solver,OPL,LeonePFEGPS06,clp}. Most modelling languages are highly
descriptive and have specific language constructs for certain sorts of constraints.
Specific constraints are translated to underlying theories by specific reduction
techniques. Some modelling systems use \sat as the target problem
\cite{GiunchigliaLM06,Huang08} and some of them focus on solving \np-complete
problems by reduction to \sat \cite{NPSpec2006}.

In this paper we present a novel approach for solving problems by reducing
them to \sat. The approach can be seen also as a general-purpose constraint
programming system (for finite domains). The approach consists of a new
specification/modelling language \ursa (from {\em Uniform Reduction to SAt})
and an associated interpreter. In contrast to other modelling languages, the
proposed language combines features of declarative and imperative programming
paradigms. What makes the language declarative is not how the constraints are
expressed, but only the fact that a procedure for finding solutions does not need
to be explicitly given. On the other hand, the system has features of imperative
languages and may be seen as an extension of the imperative programming
paradigm,\footnote{There are constraint programming libraries
for imperative languages, but these still do not follow the spirit of
imperative programming and are substantially declarative.}
similarly as some constraint programming systems are extensions of logic
programming paradigm. In contrast to other modelling languages, in the proposed
specification language loops are represented in the imperative way (not by,
generally more powerful, recursion), destructive updates are allowed, there
is support for constraints involving bitwise operators and operators for
arithmetic modulo $2^n$.
There are problems for which, thanks to these features, the modelling process
is simpler and specifications are more readable and easier to maintain than
for other languages and constraint systems. However, of course, the presented
system does not aim to replace other constraint systems and languages, but
rather to provide a new alternative with some distinctive features.

The used uniform approach enables a simple syntax and semantics of the \ursa
language, a simple, uniform reduction to \sat and, consequently, a simple
architecture of the whole system. This enables a straightforward implementation
of the proposed system and a rather straightforward verification of its
correctness. This is very important because, although it is often easier for
a declarative program than for a corresponding imperative program to verify
that it meets a given specification, this still does not lead to a high
confidence if the constraint solving system itself cannot be trusted.

The presented approach is accompanied with an open-source implementation,
publicly available on the Internet. A limited experimental comparison suggest
that the system (combined with state-of-the-art \sat solvers) yields good
performance, competitive to other modern approaches.

\paragraph{Overview of the paper.}
In Section \ref{sec:background} we give relevant definitions; in Section
\ref{sec:representation} we provide motivation and basic ideas of the proposed
approach. In Section \ref{sec:ursa_language} we describe the specification
language \ursa, in Section \ref{sec:ursa_semantics} its semantics, in Section
\ref{sec:ursa_interpreter} the corresponding interpreter, and in Section
\ref{sec:pragmatics} pragmatics of the language. In Section \ref{sec:related_work}
we discuss related techniques, languages and tools. In Section \ref{sec:future_work}
we discuss directions for future work and in Section \ref{sec:conclusions} we
draw final conclusions.

% ***************************************************************************
% ***************************************************************************
\section{Background}
\label{sec:background}
% ***************************************************************************
% ***************************************************************************

In this section we give a brief account of the \sat and \csp problems and
related notions.

\paragraph{Propositional logic.}
We assume standard notions of propositional logic: {\em literal},
{\em clause}, {\em propositional formula}, {\em conjunctive normal form}
(\cnf), {\em valuation} (or {\em assignment}), {\em interpretation},
{\em model}, {\em satisfiable formula}, etc. We denote by $\top$ and
$\bot$ the Boolean constants {\em true} and {\em false} and the logical
connectives by $\NOT$ ({\em negation}), $\AND$ ({\em conjunction}), $\OR$
({\em disjunction}), $\XOR$ ({\em exclusive disjunction}), $\IMPL$
({\em implication}), $\EQ$ ({\em equivalence}).
Two formulae $A$ and $B$ are said to be {\em equivalent} if $A$ and
$B$ have the same truth value in any valuation. Two formulae $A$ and
$B$ are said to be {\em weakly equivalent} (or {\em equisatisfiable})
if whenever $A$ is satisfiable then $B$ is satisfiable and vice versa.

\paragraph{Constraint Satisfaction Problem.}
A constraint satisfaction problem (\csp) is defined as a triple $(X,D,C)$,
where $X$ is a finite set of variables $x_1$, $x_2$, $\ldots$, $x_n$,
$D$ is a set of domains $d_1$, $d_2$, $\ldots$, $d_n$ for these variables,
and $C$ is a set of constraints $c_1$, $c_2$, $\ldots$, $c_k$. In a
finite-domain \csp, all sets from $D$ are finite. Constraints from $C$ may
define combinations of values assigned to variables that are {\em allowed}
or that are {\em prohibited}. A problem instance is satisfiable if there is
an assignment to variables such that all constraints are satisfied. Such
assignment is called a {\em solution}.
A constraint optimization problem is a \csp in which the goal is to
find a solution maximizing (or minimizing) a given {\em objective function}
over all allowed values of the given variables.

\paragraph{\sat Problem and \sat Solvers.}
\sat is the problem of deciding if a given propositional formula in \cnf
is satisfiable, i.e., if there is any assignment to variables such that all
clauses are true. Obviously, \sat is a special case of \csp, with all
variables ranging over the domain $\{ 0, 1\}$ and with constraints given
as clauses. \sat was the first problem shown to be \np-complete \cite{Cook71},
and it still holds a central position in the field of computational complexity.
Stochastic \sat solvers cannot prove the input instance to be unsatisfiable,
but may find a solution (i.e., a satisfying variable assignment) for huge
satisfiable instances quickly. On the other hand, for a given \sat instance,
a complete \sat solver always finds a satisfying variable assignment or shows
that there is no such assignment. Most of the state-of-the-art complete
\sat solvers are {\sc cdcl} (conflict-driven, clause-learning) based
extensions of the Davis-Putnam-Logemann-Loveland algorithm (\dpll)
\cite{dp,DLL62,HandbookOfSAT2009}.
In recent years, a tremendous advance has been made in \sat solving
technology \cite{MMZZM01,minisat,ZhangQuest02,HandbookOfSAT2009}.
These improvements involve both high-level and low-level algorithmic
techniques. The advances in \sat solving make possible deciding
satisfiability of some industrial \sat problems with tens of thousands
of variables and millions of clauses.

% ***************************************************************************
% ***************************************************************************
\section{Problem Specification and Problem Solving}
\label{sec:representation}
% ***************************************************************************
% ***************************************************************************

There are two basic components of the presented approach:
\begin{iteMize}{$\bullet$}
\item problem specification: a problem is specified by a test (expressed in
an imperative form) that given values of relevant variables are indeed a
solution to the problem.
\item problem solving: all relevant variables of the problem are represented
by finite vectors of propositional formulae (corresponding to vectors of bits
and to binary representation, in case of numerical values);
the specification is symbolically executed over such representation and the
assertion that given values make a solution is transformed to an instance of
the \sat problem and passed to a \sat solver. If the formula is satisfiable,
its model is transformed back to variables describing the problem, giving
a solution to the problem.
\end{iteMize}

% ***************************************************************************
\subsection{Problem Specification}
% ***************************************************************************

Let us consider problems of the following general form: {\em find
(if it exists) a set of values $S$ such that given constraints are met}
(variations of this form include: only checking if such values exists,
and finding all values that meet the given conditions).
A problem of this form can be specified by a test that checks if a given set $S$
meets the given constraints (with one assertion that combines all the constraints).
The test can be formulated in a language designed in the style
of imperative programming languages and such a test is often easy
to formulate.

\begin{exa}
\label{ex:trivial}
Let us consider a trivial problem: if $v$ equals $u+1$, find a value for $u$
such that $v$ equals $2$. A simple check in an imperative form can be specified
for this problem --- if a value of $u$ is given in advance, one could easily
check whether it is a solution of the problem. Indeed, one would assign $u+1$
to $v$ and finally check whether $v$ equals $2$. Such test can be written in
the form of an imperative C-like code (where {\tt assert(b)} checks whether
{\tt b} is true) as follows:

{\footnotesize
\begin{verbatim}
v = u+1;
assert(v==2);
\end{verbatim}
}
\end{exa}

The example above is trivial, but specifications may involve more
variables and more complex operations, including conditional
operations and loops, as illustrated by the following example.

\begin{exa}
The most popular way of generating pseudorandom numbers is based on linear
congruential generators. A generator is defined by a recurrence relation of the form:

$x_{n+1} \equiv (a x_n + c) \;\; (\mathrm{mod} \; m) \;\;\; (\mbox{for} \;\; n \geq 0)$

\noindent
and $x_0$ is the {\em seed} value ($0 \leq x_0 < m$). One example
of such relation is:

$x_{n+1} \equiv (1664525 x_n + 1013904223) \;\; (\mathrm{mod} \; 2^{32}) \;\;\; (\mbox{for} \;\; n \geq 0)$

\noindent
It is trivial to compute elements of this sequence. The check that $x_{100}$
is indeed equal to the given value if the seed is equal to $nseed$ can be
simply written in the form of an imperative C-like code as follows (assuming that
numbers are represented by $32$ bits):

{\footnotesize
\begin{verbatim}
nx=nseed;
for(ni=1;ni<=100;ni++)
  nx=nx*1664525+1013904223;
assert(nx==3998113695);
\end{verbatim}
}

\noindent
However, the following problem, realistic in simulation and testing tasks, is
a non-trivial programming problem (unless problem-specific, algebraic knowledge
is used): given, for example, the value $x_{100}$ compute $x_0$.
Still, the very same test shown above can serve as a specification of this problem.
This example illustrates one large family of problems that can be simply
specified using the proposed approach --- problems that are naturally
expressed in terms of imperative computations and that involve destructive
assignments. Such problems are often difficult to express using other
languages and systems. For the above specification, since in constraint
programming systems the destructive assignment is not allowed, in most
specification languages one would have to introduce variables for all elements
of the sequence from $x_0$ to $x_{100}$ and the constraints between any
succeeding two. Also, other systems typically do not support modular
arithmetic constraints and integers of arbitrary length.

% The \ursa system, using the above specification ($x_{100}=3998113695$)
% with the representation length $32$, generates a formula with 110231
% variables and 370730 clauses. The solution of the problem (a required
% value for \verb|nseed| --- 2011), is found in less than 2s (including
% the generating and solving phase).
\end{exa}

Note that the specifications given above also cover the information
on what variables are unknown and have to be determined so that the
constraints are satisfied ---  those are variables that appear within
commands before they were defined. So, the above code is a full and precise
specification of the problem, up to the domains of the variables. For
Boolean variables, the domain is $\{ \bot, \top \}$, while for numerical
variables a common domain interval (e.g., $[0,2^n-1]$) can be assumed and
additional domain restrictions can be given within the specification.

% ***************************************************************************
\subsection{Problem Solving}
% ***************************************************************************

The described imperative tests form problem specifications and they can be used
as a starting point in problem solving. Let us first describe a straightforward,
naive approach.

Assume that there is a problem specification (in the form of an imperative
test) and assume there is a common domain for all unknowns (except Boolean
unknowns), for example, the interval $[0,2^n-1]$ (for a given $n$).
Then, for all admissible values for all unknowns, the specification
can be executed. All sets of values satisfying the constraints should
be returned as solutions. If there are $k$ unknown numerical variables
and $l$ unknown Boolean variables, then the search space would be
of the size $2^{nk+l}$.

\begin{exa}
Let us consider the specification given in Example \ref{ex:trivial}.
If the domain for $u$ and $v$ is the interval $[0,3]$, the specification
should be executed four times and only the value $1$ assigned to $u$
leads to the constraint met, so it is the only solution of the problem.
\end{exa}

Obviously, the above naive and brute-force approach based on systematic
enumeration of all possible input values is complete (for finding all
solutions), but extremely inefficient. It can be turned to a much more
efficient version that takes into account given relationships between
variables in order to reduce the search space.
The basic idea is to represent all unknowns abstractly, in a symbolic form,
as vectors of propositional formulae. Then, all steps in the specification
can be performed using this abstracted form (i.e., can be symbolically
executed). Finally, the assertion
would generate a propositional formula for which a satisfying valuation
is to be found. If there are $k$ unknown numerical variables and $l$
unknown Boolean variables, then the number of possible valuations would
be $2^{nk+l}$ (if the interval $[0,2^n-1]$ is assumed as the domain for
numerical values). Of course, instead of a brute-force search over this
set of valuations, a \sat solver should be used (and it will typically
perform many cut-offs and search over just a part of the whole search space).

Representation of numerical variables by propositional formulae
corresponds to their binary representation. Each formula corresponds
to one bit of the binary representation. If the range of a numerical
variable is $[0,2^n-1]$, then it is represented by a vector of $n$
propositional formulae. If a bit of the number is known to be $1$,
then the corresponding formula is $\top$, and if a bit of the
number is known to be $0$, then the corresponding formula is $\bot$.
For instance, for $n=2$, $1$ is represented by $[\bot, \top]$ (where the last
position corresponds to the least significant bit). If a bit of the
number is not known, then it is represented by a propositional
variable, or, if it depends on some conditions, by a propositional
formula. We will discuss only representations of unsigned integers,
but representations of signed integers can be treated in full analogy
(moreover, floating point numbers can also be modelled in an analogous
way). Boolean variables are represented by unary vectors of propositional
formulae.

Results of arithmetic and bitwise logical operations over numbers
represented by vectors of formulae can be again represented by
propositional formulae in terms of formulae occurring in the input
arguments. If the numbers are treated as unsigned, all arithmetic
operations are performed modulo $2^n$. For instance, if $u$ is
represented by $[p,q]$ and $v$ is represented by $[r,s]$, then $v+u$
(modulo $2^2$) is represented by $[(p \XOR r) \XOR (q \AND s), q \XOR s]$.
Relational operations over numbers ($=$, $<$, $>$, $\leq$, $\geq$, $\neq$, etc.)
and logical operations and relations over Boolean values can also be
represented. For instance, if $u$ is represented by $[p,q]$ and $v$ is
represented by $[r,s]$, then $u=v$ is represented by the unary vector
$[(p \EQ r) \AND (q \EQ s)]$. Note that representations of all standard
arithmetic, Boolean, and relational operations produce polynomial
size formulae.

If a problem specification is executed over the variables represented
by vectors of propositional variables and using the corresponding
interpretation of involved operations, then the assertion of the
specification generates a propositional formula. Any satisfying valuation
(if it exists) for that formula would yield (ground) values for
numerical and Boolean unknowns that meet the specification, i.e.,
a solution to the problem.

\begin{exa}
Let us again consider Example \ref{ex:trivial}. If \verb|u| is represented
by $[p,q]$ (and \verb|1| is represented by $[\bot, \top]$), then, by the
condition \verb|v=u+1|, \verb|v| is represented by
$[(p \XOR \bot) \XOR (q \AND \top), q \XOR \top]$, i.e., after
simplification, by $[p \XOR q, \NOT q]$. From the assertion \verb|v==2|, it
follows that $[p \XOR q, \NOT q]$ should be equivalent to $[\top, \bot]$.
In other words, the formula $((p \XOR q) \EQ \top) \AND (\NOT q \EQ \bot)$
should be checked for satisfiability. It is satisfiable, and in its only
model $p$ maps to $\bot$ and $q$ maps to $\top$. Hence, the representation
for a required value of \verb|u| is $[\bot, \top]$, i.e., \verb|u| equals 1.
\end{exa}

% ***************************************************************************
\subsection{Domain of the Approach}
% ***************************************************************************

A system based on the ideas presented above, could be used not only for
combinatorial problems, but for a very wide range of problems --- it can be
used for computing $x$ such that $f(x)=y$, given $y$ and a computable function
$f$ with a finite domain and a finite range (i.e., for computing inverse of $f$).
A definition of $f$ in an imperative form can serve as a specification
of the problem in verbatim. Having such a specification of the function $f$
is a weak and realistic assumption as it is easy to make such specification
for many interesting problems, including \np-problems \cite{garey-johnson}.
If $f$ is a function such that $f(x)=1$ when $x$ is a witness for some
instance of an \np-problem, $f$ can serve as a specification for this
problem and the required answer is {\em yes} if and only if there is
$x$ such that $f(x)=1$.

Concerning the type of numbers involved, the approach can be applied
for any finite representation of signed or unsigned, integer or floating
point numbers.

In the proposed approach, all computations (over integers) are
performed modulo $2^n$. In the case of non-modular constraints, the base
can be set to a sufficiently large value.

The approach (in the presented form) cannot be used for computing $x$ such
that $f(x)=y$, for arbitrary computable function $f$. The first limitation
is a finite representation of variables. The second is that conditional
commands in the specification could involve only conditions with ground
values at the time when the condition is evaluated. However, this restriction
is not relevant for many (or most of) interesting problems.
Overall, the domain of the proposed approach covers all problems with Boolean
and numerical unknowns, over Boolean parameters and numerical parameters
with finite domains, that can be stated in the specification language that
makes the part of the approach.

% ***************************************************************************
% ***************************************************************************
\section{Syntax of \ursa Language}
\label{sec:ursa_language}
% ***************************************************************************
% ***************************************************************************

In this section we describe the language \ursa that serves as a specification
language in the spirit of the approach presented above.
A description of the syntax of the \ursa language is given, in {\sc ebnf}
representation, in Table \ref{tab:URSAsyntax} ($\langle$num var$\rangle$ denotes
the syntactical class of numerical variables, $\langle$num expr$\rangle$ denotes
the syntactical class of numerical expressions, $\langle$bool expr$\rangle$
denotes the syntactical class of Boolean expressions, etc).
An \ursa program is a sequence of statements (and procedure definitions).
There are two types of variables --- numerical, with identifiers starting
with \verb|n| and Boolean, with identifiers starting with \verb|b|. The same
convention holds for identifiers of arrays. Variables are not declared, but
introduced dynamically.
There are functions (\verb|bool2num| and \verb|num2bool|) for converting Boolean
values to numerical values and vice versa, and the \verb|sgn| function corresponding
to signum function. Arithmetic, bitwise, relational and
compound assignment operators, applied over arithmetic variables/expressions, are written
in the C-style. For example, bitwise
conjunction over numerical variables \verb|n1| and \verb|n2| is written \verb|n1 & n2|,
bitwise left shift of \verb|n1| for \verb|n2| is written \verb|n1 << n2|, and
\verb|n1 += n2| is equivalent to \verb|n1 = n1+n2|.
Logical operators, applied over Boolean  variables/expressions, are written in the
C-style, with additional operator \verb|^^| for logical exclusive disjunction, in
the spirit of other C logical operators. There are also compound assignment operators
for logical operators, such as \verb|&&=| (added for symmetry and convenience, although
they do not exist in C). The operator {\tt ite} is the conditional operator:
{\tt ite(b,n1,n2)} equals {\tt n1} if {\tt b} is true, and equals {\tt n2} otherwise.
There are no user-defined functions, but only user-defined procedures.

\begin{table}[ht!]
{\scriptsize\ttfamily
\begin{center}
\begin{tabular}{lcl}
$\langle$program$\rangle$ & ::= & $\langle$procedure def$\rangle$* $\langle$statement$\rangle$* \\ \hline

$\langle$procedure def$\rangle$ & ::= & "procedure" $\langle$procedure name$\rangle$ ( "(" ")" |  \\
                                &     & \verb| | "(" ($\langle$num var id$\rangle$ \verb_|_ $\langle$bool var id$\rangle$) ("," ($\langle$num var id$\rangle$ \verb_|_ $\langle$bool var id$\rangle$))* ")" ) \\
                                &     & \verb| | "\{" $\langle$statement$\rangle$* "\}"                                                    \\ \hline

$\langle$procedure name$\rangle$       & ::= & $\langle$letter$\rangle$($\langle$letter$\rangle$ \verb_|_ $\langle$digit$\rangle$)* \\ \hline

$\langle$statement$\rangle$ & ::= & "\{" $\langle$statement$\rangle$* "\}"    \\
            &     & \verb_|_ $\langle$num var$\rangle$ $\langle$assign num op$\rangle$ $\langle$num expr$\rangle$ ";"                       \\
            &     & \verb_|_ $\langle$num var$\rangle$ $\langle$num op postfix$\rangle$ ";" \\
            &     & \verb_|_ $\langle$bool var$\rangle$ $\langle$assign bool op$\rangle$ $\langle$bool expr$\rangle$ ";"                           \\
            &     & \verb_|_ "while" "(" $\langle$bool expr$\rangle$ ")" $\langle$statement$\rangle$                          \\
            &     & \verb_|_ "for" "(" $\langle$statement$\rangle$ ";" $\langle$bool expr$\rangle$ ";" $\langle$statement$\rangle$ ")" $\langle$statement$\rangle$            \\
            &     & \verb_|_ "if" "(" $\langle$bool expr$\rangle$ ")" $\langle$statement$\rangle$ [ "else" $\langle$statement$\rangle$ ]                          \\
            &     & \verb_|_ "call" $\langle$procedure name$\rangle$ ( "(" ")" | \\
            &     & \verb_ _ \verb| |  "(" ($\langle$num expr$\rangle$ \verb_|_ $\langle$bool expr$\rangle$) ("," ($\langle$num expr$\rangle$ \verb_|_ $\langle$bool expr$\rangle$))* ")" ) ";"  \\
            &     & \verb_|_ "minimize" "(" $\langle$num var$\rangle$ "," $\langle$num const$\rangle$ "," $\langle$num const$\rangle$ ")" ";" \\
            &     & \verb_|_ "maximize" "(" $\langle$num var$\rangle$ "," $\langle$num const$\rangle$ "," $\langle$num const$\rangle$ ")" ";" \\
            &     & \verb_|_ "assert" "(" $\langle$bool expr$\rangle$ (";" $\langle$bool expr$\rangle$)* ")" ";"                             \\
            &     & \verb_|_ "assert\_all" "(" $\langle$bool expr$\rangle$ (";" $\langle$bool expr$\rangle$)* ")" ";"                            \\
            &     & \verb_|_ "print" ($\langle$num expr$\rangle$ \verb_|_ $\langle$bool expr$\rangle$) ";"                                \\
            &     & \verb_|_ "listvars" ";"                \\
            &     & \verb_|_ "clear" ";"                   \\
            &     & \verb_|_ "halt" ";"                    \\ \hline

$\langle$num expr$\rangle$ & ::= & $\langle$num const$\rangle$                       \\
             &     &     \verb_|_ $\langle$num var$\rangle$                          \\
             &     &     \verb_|_ $\langle$un num op$\rangle$ $\langle$num expr$\rangle$  \\
             &     &     \verb_|_ $\langle$num expr$\rangle$ $\langle$num op$\rangle$ $\langle$num expr$\rangle$  \\
             &     &     \verb_|_ ite "(" $\langle$bool expr$\rangle$ "," $\langle$num expr$\rangle$ "," $\langle$num expr$\rangle$")"  \\
             &     &     \verb_|_ "sgn" "(" $\langle$num expr$\rangle$ ")"           \\
             &     &     \verb_|_ "bool2num" "(" $\langle$bool expr$\rangle$ ")"     \\
             &     &     \verb_|_ "(" $\langle$num expr$\rangle$ ")"                 \\ \hline

$\langle$num var$\rangle$ & ::= & $\langle$num var id$\rangle$ \\
             &     &     \verb_|_ $\langle$num var id$\rangle$ "[" $\langle$num expr$\rangle$ "]" \\
             &     &     \verb_|_ $\langle$num var id$\rangle$ "[" $\langle$num expr$\rangle$ "]" "[" $\langle$num expr$\rangle$ "]" \\ \hline

$\langle$num const$\rangle$ & ::= & ($\langle$digit$\rangle$)+ \\ \hline

$\langle$num var id$\rangle$ & ::= & "n"($\langle$letter$\rangle$ \verb_|_ $\langle$digit$\rangle$)* \\ \hline

$\langle$assign num op$\rangle$   & ::= & "\verb|=|" \verb_|_ "\verb|+=|" \verb_|_ "\verb|-=|" \verb_|_ "\verb|*=|" \verb_|_ "\verb|&=|" \verb_|_ "\verb_|=_" \verb_|_ "\verb|^=|" \verb_|_ "\verb|<<=|" \verb_|_ "\verb|>>=|"    \\ \hline
$\langle$num op$\rangle$          & ::= & "\verb|+|" \verb_|_ "\verb|-|" \verb_|_ "\verb|*|" \verb_|_ "\verb|&|" \verb_|_ "\verb_|_" \verb_|_ "\verb|^|" \verb_|_ "\verb|<<|" \verb_|_ "\verb|>>|"    \\ \hline
$\langle$un num op$\rangle$       & ::= & "$-$" \verb_|_ "\verb|~|"             \\ \hline
$\langle$num op postfix$\rangle$  & ::= & "\verb|++|" \verb_|_ "\verb|--|"    \\ \hline
$\langle$num rel$\rangle$         & ::= & "\verb|<|" \verb_|_ "\verb|>|" \verb_|_ "\verb|<=|" \verb_|_ "\verb|>=|" \verb_|_ "\verb|==|" \verb_|_ "\verb|!=|" \\ \hline

$\langle$bool expr$\rangle$ & ::= & $\langle$bool const$\rangle$               \\
             &     & \verb_|_ $\langle$bool var$\rangle$               \\
             &     & \verb_|_ $\langle$bool expr$\rangle$ $\langle$bool op$\rangle$ $\langle$bool expr$\rangle$            \\
             &     & \verb_|_ $\langle$un bool op$\rangle$ $\langle$bool expr$\rangle$                                        \\
             &     & \verb_|_ $\langle$num expr$\rangle$ $\langle$num rel$\rangle$ $\langle$num expr$\rangle$                  \\
             &     & \verb_|_ ite "(" $\langle$bool expr$\rangle$ "," $\langle$bool expr$\rangle$ "," $\langle$bool expr$\rangle$")"  \\
             &     & \verb_|_ "num2bool" "(" $\langle$num expr$\rangle$ ")"        \\
             &     & \verb_|_ "(" $\langle$bool expr$\rangle$ ")"                  \\ \hline

$\langle$bool const$\rangle$ & ::= & ( "true" \verb_|_ "false" ) \\ \hline

$\langle$bool var$\rangle$ & ::= & $\langle$bool var id$\rangle$ \\
               &     &     \verb_|_ $\langle$bool var id$\rangle$ "[" $\langle$num expr$\rangle$ "]" \\
               &     &     \verb_|_ $\langle$bool var id$\rangle$ "[" $\langle$num expr$\rangle$ "]" "[" $\langle$num expr$\rangle$ "]" \\ \hline

$\langle$bool var id$\rangle$ & ::= & "b"($\langle$letter$\rangle$ \verb_|_ $\langle$digit$\rangle$)* \\ \hline

$\langle$assign bool op$\rangle$ & ::= & "\verb|=|" \verb_|_ "\verb|&&=|"  \verb_|_ "\verb_||=_" \verb_|_ "\verb|^^=|"           \\ \hline
$\langle$bool op$\rangle$        & ::= & "\verb|&&|"  \verb_|_ "\verb_||_" \verb_|_ "\verb|^^|"           \\ \hline
$\langle$un bool op$\rangle$     & ::= & "\verb|!|"               \\ \hline
\end{tabular}
\end{center}
}
\caption{EBNF description of \ursa language}
\label{tab:URSAsyntax}
\end{table}

The role of the command {\tt assert} is to assert that some constraint (given as
the argument) is met (as in C). During the interpretation of the \ursa program,
this command invokes the solving mechanism and seeks for an assignment to the
introduced unknowns that make this constraint true. The command {\tt assert\_all}
is analogous, but it seeks for all satisfying models. One program can have several
commands of this sort, but each of them works locally (as in C), i.e., it invokes
the solving mechanism just for its own argument.
The instructions {\tt minimize} (and {\tt maximize}) state that a minimal (or
maximal) value within the given range for the given numerical variable should
be found. The commands {\tt assert} and {\tt assert\_all} take into account
only the last {\tt minimize}/{\tt maximize} instruction.

There are miscellaneous commands, mostly intended to be used in interactive mode
(for listing values and status of variables --- \verb|listvars|, for deleting values of
all current variables --- \verb|clear|, and stopping the interpreter --- \verb|halt|).

The following two examples illustrate some constructs of the specification language ---
the use of procedures and the use of the operator {\tt maximize}.

\begin{exa}
For a given value $k$, the task is to find all values $x$, $y$, and $z$ such that
$x^k+y^k \equiv z^k (\mathrm{mod}\; 2^n)$. The following \ursa code (with a procedure
that computes the power function) specifies the problem for $k=2$:

{\footnotesize
\begin{verbatim}
procedure power(na,nk) {
  np=na;
  for(ni=1;ni<nk;ni=ni+1)
     na = na*np;
}

nk=2;
nxpowernk=nx;
nypowernk=ny;
nzpowernk=nz;

call power(nxpowernk,nk);
call power(nypowernk,nk);
call power(nzpowernk,nk);

assert_all(nxpowernk+nypowernk==nzpowernk);
\end{verbatim}
}
\end{exa}

\begin{exa}
\label{example:maximize}
Given nine points in space, no four of which are coplanar, the task is to find the
minimal natural number $n$ such that for any coloring with red or blue of $n$ edges
drawn between these nine points there always exists a triangle having all edges of
the same color.\footnote{This is one of the problems from International Mathematical
Olympiad (IMO) held in 1992. Mathematical problems from IMOs are typically very
challenging problems from different areas of mathematics, often coming from complex
mathematical conjectures, but not requiring heavy mathematical devices themselves
(as they are aimed at high-school students) \cite{IMOcompendium}. A number of other
IMO problems can be specified and solved using the presented system.}

We will slightly reformulate the problem: it is sufficient to find the
maximal natural number $n$ such that there is a coloring with red or blue
of $n$ edges drawn between these nine points such there is no triangle
having all edges of the same color. We will assume that {\tt nE[i][j]} is
$0$ if there is no edge linking $i$-th and $j$-th point, that {\tt nE[i][j]}
equals $1$ if the edge linking $i$-th and $j$-th point is red, and that
{\tt nE[i][j]} equals $2$ if the edge linking $i$-th and $j$-th point is
blue. In the following specification, {\tt nNumberOfEdges} stores the
number of edges, {\tt bTwoColors} stores the condition that for each pair
of points, either there is no edge, or it is red or blue, and
{\tt bNoMonochromaticTriangle} stores the condition that there is no
triangle having all edges of the same color. In order to use the maximal
$n$ that meets these conditions, the command {\tt maximize(n,1,36)} is
used (there are 36 edges at most).

{\footnotesize
\begin{verbatim}
maximize(n,1,36);
nPoints=9;
nNumberOfEdges=0;
bTwoColors=true;
bNoMonochromaticTriangle=true;

for(ni=1;ni<=nPoints-1;ni++)
  for(nj=ni+1;nj<=nPoints;nj++)  {
    nNumberOfEdges += sgn(nE[ni][nj]);
    bTwoColors &&= nE[ni][nj]<3;
    for(nk=nj+1;nk<=nPoints;nk++)
      bNoMonochromaticTriangle &&=
         (nE[ni][nj]==0 || nE[ni][nj]!=nE[ni][nk] || nE[ni][nj]!=nE[nj][nk]);
  }

assert(nNumberOfEdges==n && bTwoColors && bNoMonochromaticTriangle);
\end{verbatim}
}
\end{exa}

% ***************************************************************************
\section{Semantics of \ursa Language}
\label{sec:ursa_semantics}
% ***************************************************************************

The semantics of the \ursa language is not equal, but rather parallel
to the standard semantics of imperative programming languages. Namely,
while in the standard semantics expressions (numerical and Boolean)
are always evaluated to ground values, in \ursa they may be represented
in symbolic propositional form.
In terms of operational semantics \cite{Plotkin81}, in the standard semantics,
a {\em store} (intuitively, describing memory) is a function from identifiers
to integers, while for the \ursa language, a store is a function from
identifiers to integers {\em or} vectors of propositional formulae. Boolean
variables are represented by unary vectors, while numerical variables are
represented by vectors of length $n$ (where $n$ is chosen in advance and
then fixed for one session of the interpreter's work).

A configuration is a pair $\langle c, s \rangle$ of a command or an
expression $c$ and a store $s$. A one step relation $\mapsto$ maps
configurations to configurations. In the following rule descriptions,
$\langle skip , s \rangle$ denotes a terminal configuration --- a program
that has completed execution, $x$ denotes a variable identifier, $i$, $i_1$,
and $i_2$ denote integers, $f_i$, $f'_i$, and $f''_i$ denote propositional
formulae, $e$, $e'$, $e_1$ and $e_2$ denote expressions, and
$s \ \biguplus \ (x \mapsto y)$ denotes a function $\hat{s}$ such that
$\hat{s}(x)=y$ and $\hat{s}(z)=s(z)$ for $z\neq x$.

In the standard semantics, for example, the assignment operator $=$ is
defined by the rules:
\medskip

\noindent
\begin{tabular}{ll}
(1) & $\langle x = i, s \rangle \; \mapsto \; \langle skip, s \ \biguplus \ (x \mapsto i) \rangle$ \\
(2) & $\begin{array}{ll}
\langle e, s \rangle \; \mapsto \; \langle e', s' \rangle \\ \hline
\langle x = e, s \rangle \; \mapsto \; \langle x=e' , s' \rangle
\end{array}$ \\
\end{tabular}
\medskip

Since in the \ursa language a variable can be assigned both a ground
integer value or a vector of propositional formulae, the semantics of
the assignment operator in \ursa is defined by the rules (the rule
(1) from above is split into two rules):

\medskip
\noindent
\begin{tabular}{ll}
(1'a) & $\langle x = i, s \rangle \; \mapsto \; \langle skip, s \ \biguplus \ (x \mapsto i) \rangle$ \\
(1'b) & $\langle x = [f_1, \ldots, f_n], s \rangle \; \mapsto \; \langle skip, s \ \biguplus \ (x \mapsto [f_1, \ldots, f_n]) \rangle$ \\
(2') & $\begin{array}{ll}
\langle e, s \rangle \; \mapsto \; \langle e', s' \rangle \\ \hline
\langle x = e, s \rangle \; \mapsto \; \langle x=e' , s' \rangle
\end{array}$ \\
\end{tabular}
\medskip

If, in an assignment command, the right hand side is evaluated to a
ground integer (i.e., if the rule (1'a) has been applied), then the
variable on the left hand side gets the status {\em ground}.
If the right hand side is evaluated to a vector of formulae (i.e.,
if the rule (1'b) has been applied), then the variable on the left
hand side gets the status {\em symbolic} and {\em dependent}.

In the standard semantics, for an operator $\diamond$ over integers,
the relation $\mapsto$ is defined for expressions by the following
four rules:
\medskip

\noindent
\begin{tabular}{ll}
(3) & $\langle x, s \rangle \; \mapsto \; \langle s(x), s \rangle$ \\

(4) & $\langle i_1 \diamond i_2, s \rangle \; \mapsto \; \langle i, s \rangle$ (where $i$ equals $i_1 \diamond i_2$) \\

(5) & $\begin{array}{ll}
\langle e_1, s \rangle \; \mapsto \; \langle e'_1, s' \rangle \\ \hline
\langle e_1 \diamond e_2, s \rangle \; \mapsto \; \langle e'_1 \diamond e_2, s' \rangle
\end{array}$ \\

(6) & $\begin{array}{ll}
\langle e_2, s \rangle \; \mapsto \; \langle e'_2, s' \rangle \\ \hline
\langle i \diamond e_2, s \rangle \; \mapsto \; \langle i \diamond e'_2, s' \rangle
\end{array}$ \\
\end{tabular}
\medskip

The corresponding rules for the \ursa language are substantially different.
Regarding the rule (3), in standard imperative programming languages, a
statement attempting to evaluate an undefined variable (one that is not
in the domain of the current store) results in a runtime error. On the
other hand, in the \ursa language, undefined variables can be used in
expressions, so this introduces a non-standard rule (instead of the
rule (3))\footnote{More precisely, rather than a rule, this is a rule-schema,
with instances for each $n$.}:
\medskip

\noindent
\begin{tabular}{ll}
(3') &
$\langle x , s \rangle \; \mapsto \;
\left\{\begin{array}{ll}
\langle s(x) , s \rangle & \mbox{if $x$ is defined in $s$} \\
\langle [v_1, \ldots, v_n] , s \ \biguplus \ (x \mapsto [v_1, \ldots, v_n]) \rangle & \mbox{otherwise ($v_i$'s are fresh} \\
                                                                                    & \mbox{propositional variables)}
\end{array}\right.
$
\end{tabular}
\medskip

If, for a variable $x$, the second case of the above rule has been applied,
$x$ gets the status {\em symbolic} and {\em independent} (and is
represented by a vector of propositional formulae). In other words,
this status is associated to each variable that is for the first
time used not on the left hand side of an assignment operator.
For example, if a variable \verb|nX| was not defined beforehand, after the
statement \verb|nY=nX;|, both variables are to be represented by the same
sequence of propositional variables, but \verb|nX| will internally have the
status {\em independent}, while \verb|nY| will have the status {\em dependent}.
Variables with the status {\em symbolic} and {\em independent} are those that should
be determined in order to solve the given problem.

\begin{exa}
Let us consider the specification for the trivial problem from
Example \ref{ex:trivial}:

{\footnotesize
\begin{verbatim}
nv = nu+1;
assert(nv==2);
\end{verbatim}
}

The variable {\tt nu} is used for the first time within the first command, so
it is symbolic (internally represented by a vector of propositional formulae)
and independent. The variable {\tt nv} becomes symbolic and dependent (it depends
on {\tt nu}).
\end{exa}

Since in \ursa expressions (numerical and Boolean) may be represented
in symbolic propositional form, the rule (4) is replaced by the
following four rules:
\medskip

\noindent
\begin{tabular}{ll}
(4'a) & $\langle i_1 \diamond i_2, s \rangle \; \mapsto \; \langle i, s \rangle$ (where $i$ equals $i_1 \diamond i_2$) \\

(4'b) & $\langle i \diamond [f'_1, \ldots, f'_n], s \rangle \; \mapsto \; \langle [f_1, \ldots, f_n] \diamond [f'_1, \ldots, f'_n], s \rangle$ \\
      & (where $[f_1, \ldots, f_n]$ is a binary representation of $i$) \\

(4'c) & $\langle [f_1, \ldots, f_n] \diamond i, s \rangle \; \mapsto \; \langle [f_1, \ldots, f_n] \diamond [f'_1, \ldots, f'_n], s \rangle$ \\
      & (where $[f'_1, \ldots, f'_n]$ is a binary representation of $i$) \\

(4'd) & $\langle [f_1, \ldots, f_n] \diamond [f'_1, \ldots, f'_n], s \rangle \; \mapsto \; \langle [f''_1, \ldots, f''_n], s \rangle$
\end{tabular}
\medskip

\noindent
where, in the last rule, $[f''_1, \ldots, f''_n]$ is a vector of propositional
formulae corresponding to $[f_1, \ldots, f_n] \diamond [f'_1, \ldots, f'_n]$,
assuming that all numerical expressions are considered unsigned\footnote{
Dealing with representation of signed integers or floating point numbers
can be, in principle, described and implemented by analogy
(\url{http://www.informatik.uni-bremen.de/~florian/sonolar/},
\url{http://www.cprover.org/SMT-LIB-Float/} \cite{Lerda,BrilloutKW09}). Still,
as discussed by Brillout et al.~\cite{BrilloutKW09}, the bottleneck of a reducing
floating point operations to propositional logic would be in the complexity of the
resulting propositional formulae.}
and all arithmetic operations are performed modulo $2^n$. For example,
assuming that numbers are represented by vectors of length 2 (i.e., $n=2$),
the rule (4'd) for the operator $+$ is as follows:
\medskip

\begin{tabular}{ll}
   &
$\langle [f_1,f_2] + [f'_1,f'_2], s \rangle \; \mapsto \;
    \langle [(f_1 \XOR f'_1) \XOR (f_2 \AND f'_2), f_2 \XOR f'_2] , s \rangle$
\end{tabular}
\medskip

As another example, assuming that numbers are represented by vectors of
length 2 (i.e., $n=2$), the operator $>$ is defined as follows
(in accordance with lexicographic ordering):
\medskip

\begin{tabular}{ll}
   &
$\langle [f_1,f_2] > [f'_1,f'_2], s \rangle \; \mapsto \;
    \langle [(f_1 \AND \NOT f'_1) \OR ((f_1 \EQ f'_1) \AND (f_2 \AND \NOT f'_2)] , s \rangle$
\end{tabular}
\medskip

By the above semantics, an expression involving symbolic expressions is
also a symbolic expression. However, for efficiency reasons, if it is
possible, an expression will be computed to a ground value. For instance,
after the statement \verb|nA = nB * 0;|, the variable \verb|nA| will have a
value \verb|0| and after the statement \verb|bX = bY && false;|, the variable
\verb|bX| will have the ground value {\em false}, even if the variables
\verb|bB| and \verb|bY| were symbolic. This is described by the modified
version of the rule (4'd):
\medskip

\noindent
\begin{tabular}{ll}
(4''d) &
$\langle [f_1, \ldots, f_n] \diamond [f'_1, \ldots, f'_n], s \rangle \; \mapsto \;
\left\{\begin{array}{ll}
\langle i, s \rangle & \mbox{\footnotesize{if all $f''_1$, $\ldots$, $f''_n$ are Boolean}} \\
                     & \mbox{\footnotesize{constants, where $i$ is a number}} \\
                     & \mbox{\footnotesize{with the binary representation}} \\
                     & \mbox{\footnotesize{$[f''_1, \ldots, f''_n]$}} \\
\langle [f''_1, \ldots, f''_n] , s \rangle & \mbox{\footnotesize{otherwise}}
\end{array}\right.$
\end{tabular}
\medskip

While the rule (5) from above is kept unchanged for the \ursa language,
the rule (6) is split into two rules, one for ground argument and one
for symbolic argument:
\medskip

\noindent
\begin{tabular}{ll}
(5') & $\begin{array}{ll}
\langle e_1, s \rangle \; \mapsto \; \langle e'_1, s' \rangle \\ \hline
\langle e_1 \diamond e_2, s \rangle \; \mapsto \; \langle e'_1 \diamond e_2, s' \rangle
\end{array}$ \\

(6'a) & $\begin{array}{ll}
\langle e_2, s \rangle \; \mapsto \; \langle e'_2, s' \rangle \\ \hline
\langle i \diamond e_2, s \rangle \; \mapsto \; \langle i \diamond e'_2, s' \rangle
\end{array}$ \\

(6'b) & $\begin{array}{ll}
\langle e_2, s \rangle \; \mapsto \; \langle e'_2, s' \rangle \\ \hline
\langle [f_1, \ldots, f_n] \diamond e_2, s \rangle \; \mapsto \; \langle [f_1, \ldots, f_n] \diamond e'_2, s' \rangle
\end{array}$
\end{tabular}
\medskip

The semantics of \verb|while| is defined by the following standard rules:

\noindent
\begin{tabular}{ll}
(7) &
$\begin{array}{ll}
\langle b , s \rangle \; \mapsto \; \langle \top , s' \rangle \\ \hline
\langle \mathtt{while} \;\; b \;\; \mathtt{do} \;\; c, s \rangle \; \mapsto \; \langle c ; \mathtt{while}\;\; b \;\; \mathtt{do}\;\; c, s' \rangle
\end{array}$ \\

(8) &
$\begin{array}{ll}
\langle b , s \rangle \; \mapsto \; \langle \bot , s' \rangle \\ \hline
\langle \mathtt{while} \;\; b \;\; \mathtt{do} \;\; c, s \rangle \; \mapsto \; \langle skip, s' \rangle
\end{array}$ \\
\end{tabular}
\medskip

If, for the statement $\mathtt{while} \;\; b \;\; \mathtt{do} \;\; c$,
neither of ${\langle b , s \rangle \; \mapsto \; \top}$ and
${\langle b , s \rangle \; \mapsto \; \bot}$ holds (i.e., if $b$ is evaluated
to a propositional formulae), then a run-time error is raised. The semantics
of \verb|for| is defined by analogy. This is a restriction of the specification
language and it is difficult to overcome, as it would require indefinite loop
unrolling. Recursion is also not supported and an index for accessing an
array element has to be a ground number. A condition for the \verb|if| statement
has to evaluate to a ground Boolean value, too (otherwise a run-time error is
raised). However, as a substitute, there is a conditional statement $\mathtt{ite}$
(for {\em if-then-else}) and its condition (the first argument) can be either
ground or symbolic. Its semantics, if the last two arguments are of the Boolean
type, is defined in the following way (if the arguments are of the numerical
type, it is defined for each vector element by analogy):
\medskip

\begin{tabular}{ll}
   &
$\langle \mathtt{ite}(b,[b_1],[b_2], s \rangle \; \mapsto \;
    \langle (b \IMPL b_1) \AND (\NOT b \IMPL b_2) , s \rangle$
\end{tabular}
\medskip

Arguments to procedures are passed by name if they are variables, and
otherwise, they are passed by value.

The semantics of other commands in the \ursa language and its relationship
with the standard semantics is obvious or analogous to the semantics of
the commands given above.

The above, clear and relatively simple semantics, enable a rather straightforward
(although still tedious) verification (of correctness) of the proposed system:
if the system returns a solution, then this solution indeed meets the given
specification, and --- if the system does not return a solution, then the
specification has no solutions. The correctness property is given by the
following theorem ($s_\emptyset$ denotes a function not defined for any argument
and $\mapsto$ correspond both to the standard and the \ursa semantics since there
are no unknowns in the relevant specification).

\begin{thm}
If the variables $v_1$, $v_2$, $\ldots$, $v_n$ are (the only) unknowns in an \ursa
specification $S;assert(b);$, then it leads to a solution $(v_1,v_2,\ldots,v_n)=(c_1,c_2,\ldots,c_n)$,
(i.e. this solution leads to the model of the formula $s(b)$)
iff $\langle v_1=c_1;v_2=c_2;\ldots;v_n=c_n;S;assert(b), s_\emptyset \rangle$
$\mapsto$ $\langle skip, s \rangle$ where $s(b)$ is $true$.\qed
\end{thm}

The above semantics also ensure {\em faithfulness} \ of problem specifications ---
each model meeting the specification leads to one model of the generated
propositional formula, and each model of the generated propositional
formula leads to one model of the input specification. This property
is essential for constructing (or counting) all solutions of a problem.

% ***************************************************************************
% ***************************************************************************
\section{Interpreter for \ursa Language}
\label{sec:ursa_interpreter}
% ***************************************************************************
% ***************************************************************************

In this section we describe the implementation of the interpreter
for the \ursa language.\footnote{The source code of the interpreter and example
specifications are available within the distribution of the \ursa tool,
available online from: \url{http://www.matf.bg.ac.rs/~janicic/ursa.zip}.}

The interpreter is implemented in the programming language C++. The
whole system has a simple architecture and is relatively small (around
100Kb of source code, not counting \sat solver used). The overall
architecture of the system is illustrated in Figure \ref{fig:architecture}.

\begin{figure}
\begin{center}
\framebox[12cm][c]{Parsing and interpreting \ursa statements} \\
$\downarrow$ \\
\framebox[12cm][c]{Generating a goal formula from the constraint} \\
$\downarrow$ \\
\framebox[12cm][c]{Transforming the goal formula to \cnf} \\
$\downarrow$ \\
\framebox[12cm][c]{Invoking the \sat solver} \\
$\downarrow$ \\
\framebox[12cm][c]{Turning \sat models into values of the independent variables}
\end{center}
\caption{Overall architecture of \ursa system}
\label{fig:architecture}
\end{figure}

\subsection{Parsing and Interpreting \ursa Statements}
\label{subsec:parsing}

Statements can be entered in an interactive mode and processed one
by one or provided to the \ursa system within a file. Statements are
interpreted according to the semantics described in Section
\ref{sec:ursa_semantics}.

The table of symbols (i.e., the {\em store}) stores values of numerical
and Boolean variables. Numerical variables are represented by a class
(\verb|Number|) whose objects can be either symbolic numbers (i.e., vectors
of propositional formulae) or concrete numbers. For this class all standard
arithmetic (except division\footnote{It is possible to define division
in symbolic terms, but for its complexity, it is not implemented
in the current version of the system. Still, division can be
modelled within the system by using multiplication as shown in the following
example, giving {\tt nD} as a results of integer
division {\tt nX/nY} and {\tt nR} as a remainder (the constraint on
{\tt nD} restricts the number of solutions, because of the modular arithmetic):
{\tt assert(nX==nY*nD+nR;nR<nY;nD<nX);}}), bitwise logical, and
relational operators are defined. In principle, these operators return ground
values for ground operands and vectors of formulae for symbolic operands (but,
if possible, return ground values for symbolic operands). Boolean variables
are represented by analogy (by the class \verb|Boolean|).

Symbolic values of variables are represented by vectors of propositional
formulae. There are classes for dealing with propositional formulae
(\verb|Formula|) and vectors of propositional formulae (\verb|FormulaVector|).
The length of vectors of propositional formulae representing numerical
variables is given as a parameter to the interpreter (the default value is 8,
corresponding to integers from 0 to 255).

In the current implementation, all numerical variables and constants are
treated as unsigned integers (represented by binary representation).
For the class \verb|FormulaVector| all standard arithmetic (except division),
bitwise logical, and relational operators are defined. They deal only
with symbolic numerical and Boolean values (in contrast to the class
\verb|Number| and \verb|Boolean|). The implementations of bitwise
logical operators are simple and straightforward. The implementations
of the arithmetic operators modulo $2^n$ and the relational operators
are more complex because resulting bits depend on all the previous
bits of the operands. Figure \ref{fig:greater_implementation}
shows the implementation of the relational operator $>$ (it processes
propositional formulae corresponding to bits from the least significant
one and returns a unary vector).
Note that the given function is not computing a value for the
relational operator $>$ for concrete numbers, but it sets a formula that
corresponds to $>$ for two symbolic input numerical expressions.

\begin{figure}[ht]
{\footnotesize
\begin{verbatim}
FormulaVector1 FormulaVector::operator > (const FormulaVector &fv) {
  FormulaVector1 result, r1,r2;
  r1 = (*this)[size-1];
  r2 = fv[size-1];
  result = (r1 & ~r2);
  for(int i=size-2;i>=0;i--) {
    r1 = (*this)[i];
    r2 = fv[i];
    result = (r1 & ~r2) | (result & (r1==r2));
  }
  return result;
}
\end{verbatim}
}
\caption{Implementation of the operator $>$.}
\label{fig:greater_implementation}
\end{figure}

For the sake of efficiency (both time and space), the technique of
shared expressions was used. So, each subformula is stored only
once, but there can be more references to it (from different formulae).
All formulae generated during the interpretation of the program
are stored by the class \verb|FormulaFactory| in this way.
Hence, the formulae are not stored individually, but in a form
of a directed graph that stores links between them. Each formula
that is not a propositional variable is represented by its
connective and by references to its subformulae. Each formula is
assigned a unique (numerical) identifier.

\begin{exa}
Figure \ref{fig:formulaDAG} illustrates how the (artificial) formula
$(p\AND{}(q\AND{}r))\OR((q\AND{}r)\AND{}\NOT{}p)$ and its subformulae
are stored internally.

\begin{figure}
\begin{center}
\begin{tikzpicture}
\clip (0,0) rectangle (11.000000,4.400000);
{\footnotesize

% Marking point p by circle
\draw [line width=0.016cm] (1.000000,0.600000) circle (0.180000);%
\draw (1.110000,0.490000) node [anchor=north west] { $p$ };%

% Marking point q by circle
\draw [line width=0.016cm] (5.000000,0.600000) circle (0.180000);%
\draw (5.110000,0.490000) node [anchor=north west] { $q$ };%

% Marking point r by circle
\draw [line width=0.016cm] (9.000000,0.600000) circle (0.180000);%
\draw (9.110000,0.490000) node [anchor=north west] { $r$ };%

% Printing text at point p
\draw (1.000000,0.600000) node  { $1$ };%

% Printing text at point q
\draw (5.000000,0.600000) node  { $2$ };%

% Printing text at point r
\draw (9.000000,0.600000) node  { $3$ };%

% Marking point q\AND{}r by circle
\draw [line width=0.016cm] (4.000000,1.800000) circle (0.180000);%
\draw (4.200000,1.800000) node [anchor=west] { $q\AND{}r$ };%

% Printing text at point q\AND{}r
\draw (4.000000,1.800000) node  { $4$ };%

% Printing text at point q\AND{}r
\draw (3.800000,1.800000) node [anchor=east] { $\AND$ };%

% Drawing vector A M
\draw [line width=0.016cm] (4.115233,1.661720) -- (4.750000,0.900000);%
\draw [line width=0.016cm] (4.589670,1.153563) -- (4.750000,0.900000);%
\draw [line width=0.016cm] (4.589670,1.153563) -- (4.685982,0.976822);%
\draw [line width=0.016cm] (4.529506,1.103426) -- (4.750000,0.900000);%
\draw [line width=0.016cm] (4.529506,1.103426) -- (4.685982,0.976822);%

% Drawing segment M B
\draw [line width=0.016cm] (4.750000,0.900000) -- (4.884767,0.738280);%

% Drawing vector A M
\draw [line width=0.016cm] (4.175030,1.757993) -- (7.750000,0.900000);%
\draw [line width=0.016cm] (7.469918,1.007490) -- (7.750000,0.900000);%
\draw [line width=0.016cm] (7.469918,1.007490) -- (7.652761,0.923337);%
\draw [line width=0.016cm] (7.451641,0.931336) -- (7.750000,0.900000);%
\draw [line width=0.016cm] (7.451641,0.931336) -- (7.652761,0.923337);%

% Drawing segment M B
\draw [line width=0.016cm] (7.750000,0.900000) -- (8.824970,0.642007);%

% Marking point \NOT{}p by circle
\draw [line width=0.016cm] (8.000000,1.800000) circle (0.180000);%
\draw (8.200000,1.800000) node [anchor=west] { $\NOT{}p$ };%

% Printing text at point \NOT{}p
\draw (8.000000,1.800000) node  { $5$ };%

% Printing text at point \NOT{}p
\draw (7.800000,1.800000) node [anchor=east] { $\NOT$ };%

% Drawing vector A M
\draw [line width=0.016cm] (7.822588,1.769587) -- (2.750000,0.900000);%
\draw [line width=0.016cm] (3.049773,0.911661) -- (2.750000,0.900000);%
\draw [line width=0.016cm] (3.049773,0.911661) -- (2.848562,0.916896);%
\draw [line width=0.016cm] (3.036541,0.988850) -- (2.750000,0.900000);%
\draw [line width=0.016cm] (3.036541,0.988850) -- (2.848562,0.916896);%

% Drawing segment M B
\draw [line width=0.016cm] (2.750000,0.900000) -- (1.177412,0.630413);%

% Marking point p\AND{}(q\AND{}r) by circle
\draw [line width=0.016cm] (4.000000,3.000000) circle (0.180000);%
\draw (4.200000,3.000000) node [anchor=west] { $p\AND{}(q\AND{}r)$ };%

% Printing text at point p\AND{}(q\AND{}r)
\draw (4.000000,3.000000) node  { $6$ };%

% Printing text at point p\AND{}(q\AND{}r)
\draw (3.800000,3.000000) node [anchor=east] { $\AND$ };%

% Marking point (q\AND{}r)\AND{}\NOT{}p by circle
\draw [line width=0.016cm] (8.000000,3.000000) circle (0.180000);%
\draw (8.200000,3.000000) node [anchor=west] { $(q\AND{}r)\AND{}\NOT{}p$ };%

% Printing text at point (q\AND{}r)\AND{}\NOT{}p
\draw (8.000000,3.000000) node  { $7$ };%

% Printing text at point (q\AND{}r)\AND{}\NOT{}p
\draw (7.800000,3.000000) node [anchor=east] { $\AND$ };%

% Drawing vector A M
\draw [line width=0.016cm] (4.000000,2.820000) -- (4.000000,2.100000);%
\draw [line width=0.016cm] (4.039158,2.397433) -- (4.000000,2.100000);%
\draw [line width=0.016cm] (4.039158,2.397433) -- (4.000000,2.200000);%
\draw [line width=0.016cm] (3.960842,2.397433) -- (4.000000,2.100000);%
\draw [line width=0.016cm] (3.960842,2.397433) -- (4.000000,2.200000);%

% Drawing segment M B
\draw [line width=0.016cm] (4.000000,2.100000) -- (4.000000,1.980000);%

% Drawing vector A M
\draw [line width=0.016cm] (7.827591,2.948277) -- (5.000000,2.100000);%
\draw [line width=0.016cm] (5.296142,2.147960) -- (5.000000,2.100000);%
\draw [line width=0.016cm] (5.296142,2.147960) -- (5.095783,2.128735);%
\draw [line width=0.016cm] (5.273638,2.222973) -- (5.000000,2.100000);%
\draw [line width=0.016cm] (5.273638,2.222973) -- (5.095783,2.128735);%

% Drawing segment M B
\draw [line width=0.016cm] (5.000000,2.100000) -- (4.172409,1.851723);%

% Drawing vector A M
\draw [line width=0.016cm] (3.859444,2.887555) -- (1.750000,1.200000);%
\draw [line width=0.016cm] (2.006718,1.355228) -- (1.750000,1.200000);%
\draw [line width=0.016cm] (2.006718,1.355228) -- (1.828087,1.262470);%
\draw [line width=0.016cm] (1.957795,1.416382) -- (1.750000,1.200000);%
\draw [line width=0.016cm] (1.957795,1.416382) -- (1.828087,1.262470);%

% Drawing segment M B
\draw [line width=0.016cm] (1.750000,1.200000) -- (1.140556,0.712445);%

% Drawing vector A M
\draw [line width=0.016cm] (8.000000,2.820000) -- (8.000000,2.100000);%
\draw [line width=0.016cm] (8.039158,2.397433) -- (8.000000,2.100000);%
\draw [line width=0.016cm] (8.039158,2.397433) -- (8.000000,2.200000);%
\draw [line width=0.016cm] (7.960842,2.397433) -- (8.000000,2.100000);%
\draw [line width=0.016cm] (7.960842,2.397433) -- (8.000000,2.200000);%

% Drawing segment M B
\draw [line width=0.016cm] (8.000000,2.100000) -- (8.000000,1.980000);%

% Marking point (p\AND{}(q\AND{}r))\OR((q\AND{}r)\AND{}\NOT{}p) by circle
\draw [line width=0.016cm] (6.000000,4.200000) circle (0.180000);%
\draw (6.200000,4.200000) node [anchor=west] { $(p\AND{}(q\AND{}r))\OR((q\AND{}r)\AND{}\NOT{}p)$ };%

% Printing text at point (p\AND{}(q\AND{}r))\OR((q\AND{}r)\AND{}\NOT{}p)
\draw (6.000000,4.200000) node  { $8$ };%

% Printing text at point (p\AND{}(q\AND{}r))\OR((q\AND{}r)\AND{}\NOT{}p)
\draw (5.800000,4.200000) node [anchor=east] { $\OR$ };%

% Drawing vector A M
\draw [line width=0.016cm] (5.845651,4.107391) -- (4.500000,3.300000);%
\draw [line width=0.016cm] (4.775194,3.419451) -- (4.500000,3.300000);%
\draw [line width=0.016cm] (4.775194,3.419451) -- (4.585749,3.351450);%
\draw [line width=0.016cm] (4.734901,3.486606) -- (4.500000,3.300000);%
\draw [line width=0.016cm] (4.734901,3.486606) -- (4.585749,3.351450);%

% Drawing segment M B
\draw [line width=0.016cm] (4.500000,3.300000) -- (4.154349,3.092609);%

% Drawing vector A M
\draw [line width=0.016cm] (6.154349,4.107391) -- (7.500000,3.300000);%
\draw [line width=0.016cm] (7.265099,3.486606) -- (7.500000,3.300000);%
\draw [line width=0.016cm] (7.265099,3.486606) -- (7.414251,3.351450);%
\draw [line width=0.016cm] (7.224806,3.419451) -- (7.500000,3.300000);%
\draw [line width=0.016cm] (7.224806,3.419451) -- (7.414251,3.351450);%

% Drawing segment M B
\draw [line width=0.016cm] (7.500000,3.300000) -- (7.845651,3.092609);%
}
\end{tikzpicture}
\caption{Internal representation of formulae} \label{fig:formulaDAG}
\end{center}
\end{figure}
\end{exa}

% ***************************************************************************
\subsection{Generating a Goal Formula from the Constraint}
\label{subsec:translation_to_sat}
% ***************************************************************************

When executing the command \verb|assert|, the interpreter invokes the
underlying solving process.
For the given constraint (given as an argument to
{\tt assert}), a corresponding (single) propositional formula is
generated, transformed to \cnf and, by a \sat solver, it is
checked whether the formula is true in some valuation. If yes, then the
system lists values of all independent variables in that valuation. The
command \verb|assert_all| is similar, but it lists values of independent
variables in all satisfying valuations.

Transforming the formula $F$ that corresponds to the constraint to \cnf
is performed using Tseitin transformation \cite{Tseitin68} which is linear
in both space and time. The central idea of the transformation is to
introduce new (,,definitional``) variables for all subformulae of the
input formula, to replace the subformulae with the corresponding variables,
and to add conjuncts (in the form of clauses) representing ,,definitions``
of newly introduced variables. This yields a resulting formula that is
not logically equivalent to $F$ (because of the newly introduced variables)
but is {\em equisatisfiable} to it (i.e., the resulting formula is
satisfiable if and only if $F$ is satisfiable). Still, each model for the
resulting formula gives one model for $F$ and vice versa (thanks to the
nature of definitional variables).
This approach is suitable for the representation of formulae
described in Section \ref{subsec:parsing} since all subformulae of the
formula representing the constraint are already generated and assigned
unique identifiers that correspond to definitional variables.

The problem with the Tseitin transformation is that it introduces
many new variables, and consequently many clauses.
There are techniques that can reduce the number of variables and clauses,
e.g., by using implications instead of equivalences for subformulae that
occur in one polarity only \cite{Egly96}.
In the current implementation of the \ursa system, for reducing (in some
cases) the number of variables and clauses, associativity and commutativity
of the connectives $\AND$ and $\OR$ is used.\footnote{
The fact that the connectives $\AND$ and $\OR$ are associative and commutative
is used only when the formula is already generated. An alternative would be to
deal with n-ary conjunctions and disjunctions in earlier stages, along the
generation process.}
If a formula $A$ is of the form $A_1 \AND A_2 \AND \ldots \AND A_n$ there is
no need to introduce new variables for each of $n-1$ conjunctions. Indeed,
the standard Tseitin set of resulting clauses for this case can be replaced by:
$$(p_A \OR \NOT p_{A_1} \OR \NOT p_{A_2} \OR \ldots \OR \NOT p_{A_n})
\AND (\NOT p_A \OR p_{A_1}) \AND (\NOT p_A \OR p_{A_2}) \AND \ldots (\NOT p_A \OR p_{A_n})$$
Disjunctions are treated by analogy.
In addition, if the initial formula is a conjunction, its conjuncts are put
into \cnf separately and conjoined. If any of those conjuncts is a clause,
it is directly used (without transformation and definitional variables) at
the conjunction top level \cite{Harrison}. The above modifications still keep
the transformation linear.

Thanks to the above modifications, input instances of the \sat problem itself
(represented simply, in a way that is close to the \dimacs format)
are reduced by the \ursa system to the same instances, up to renaming
variables.\footnote{
Because of the possible renaming of the variables, solving the same \sat
instance by \ursa and by the underlying solver would not necessarily
take the same time as the solving process can take different routes.}

Typically, reducing the size of a formula is an important objective, but a
smaller number of clauses does not necessarily mean an easier formula and
sometimes adding the right clauses (e.g., those corresponding to {\em symmetry
breaking}) can have a positive effect to performance of the solver. Still,
guessing and adding clauses that could speed-up the solving process is
is not performed in the current version of \ursa.

% ***************************************************************************
\subsection{Invoking the \sat Solver}
% ***************************************************************************

The \ursa system (currently) uses state-of-the-art \sat solvers
\argosat\footnote{\argosat (\url{http://argo.matf.bg.ac.rs/downloads.html})
is an open-source, flexible, and verified \sat solver \cite{dpll-correctness,jar-sat-correctness}.}
and {\sc clasp}.\footnote{{\sc clasp} (\url{http://www.cs.uni-potsdam.de/clasp/})
is a solver for (extended) normal logic programs \cite{GebserKNS07a}. It combines
high-level modelling capacities of {\sc asp} with state-of-the-art techniques from
the area of \sat solving and it can be used as an {\sc asp} solver or a {\sc sat}
solver. {\sc clasp} was a winner at the {\sc asp} Competition 2009 in several
categories (\url{http://www.cs.kuleuven.be/~dtai/events/ASP-competition/Results.shtml})
and a winner at the \sat Competition 2009 (\url{http://www.satcompetition.org/})
in categories crafted \sat+{\sc unsat} and Crafted \sat.}
The solvers are called directly through appropriate function calls.
Alternatively, the solvers could accept inputs through files in the
standard \dimacs form\footnote{In \dimacs form, the number of variables $N$
and the number of clauses $L$ are given first, and are followed by the
list of clauses. The variables are represented by natural numbers (from
$1$ to $N$) and their negations are represented by corresponding negative
numbers (from $-1$ to $-N$). More details about the \dimacs format can
be found online: \url{http://www.satlib.org/Benchmarks/SAT/satformat.ps}.}
(\ursa can be used just to translate the input problem to \sat, without solving
it). In that case, the solvers can be used as a black-box, and could be replaced
by any \sat solver that accepts this input format (and has an option to generate
all models of the input propositional formula).

Since during the transformation to \cnf (described above) some variables
are eliminated, it is necessary to rename all remaining variables
(and update the numbers of variables). Namely, if some variables do
not occur in the generated \cnf, the solver would consider as different
all models that differ only in the values of such variables. The basic,
independent variables (those that correspond to independent \ursa variables)
are never eliminated, even if they don't occur in the generated formula
(because their values in all models are relevant).

Along with the generated \sat instance, \ursa can also pass to the solver
the information on which propositional variables are independent and which
are dependent. Some \sat solvers can use this information to guide
(and hopefully speed-up) the solving process.
If the constraint is given by the \verb|assert_all| command, the solver is
invoked to generate all satisfying valuations.\footnote{If the solver does
not have a feature for generating all satisfying valuations, a simple approach
with {\em blocking clauses} and successive calls to the solver can be used
(a blocking clause, added to the set of initial clauses, consists of all
literals defining one model negated --- this way, a model once generated
will not be repeated.}
When multiple invocations of the solver are used (for finding all solutions),
the solving process can benefit from clauses learnt in previous invocations.

If there is a statement \verb|minimize| (or \verb|maximize|) used, in the current
implementation, the problem is solved sequentially for all values from the given
range assigned to a relevant variable --- from the minimal (maximal) element in
the range onwards, seeking for a minimal (maximal) value that meets the constraint.

% ***************************************************************************
\subsection{Turning \sat Models into Values of the Independent Variables}
% ***************************************************************************

If the \sat solver finds a satisfying valuation for the input formula,
that valuation is used for computing values of independent \ursa variables.
For this, only the basic variables are relevant (and not the definitional
variables introduced during transformation to \cnf). Each satisfying
valuation determines a vector of fixed Boolean values that correspond
to an independent variable. The numerical (and Boolean) values are trivially
computed from such representations and returned by the \ursa system.

% ***************************************************************************
% ***************************************************************************
\section{Pragmatics of \ursa Language}
\label{sec:pragmatics}
% ***************************************************************************
% ***************************************************************************

Representation of symbolic values natively used in \ursa corresponds to
binary representation of unsigned integers, but the specification language
is expressive enough and leaves enough freedom for modelling problems in
substantially different ways and in different encoding styles, leading to
simpler or more efficient solutions.
For illustrating this, a prototypical \csp example --- the queens problem is
considered. The problem is to place $N$ chess queens on an $N \times N$ chessboard
such that none of them is able to capture any other (following the standard chess
queen's moves).
For modelling this problem, in the following text we do not use symmetry breaking
or other similar additional constraints, but only the basic formulations of
the problem.

Since each row (denoted by numbers $0$ to $N-1$) of the board can have exactly
one queen on it, the problem is, simply reformulated, to determine, for each row,
a column for one queen to be placed. Hence, a solution to the problem is a
sequence $r_0$, $r_1$, $\ldots$, $r_{N-1}$ that meet the given constraints.
If such sequence is given in advance, it can be simply checked whether it is
indeed a solution of the problem. The next \ursa program, followed by a fragment
of the output, specifies the problem (for the problem size $8$; in this and other
\ursa specifications that follow, the dimension of an instance can be trivially
changed in one line).

{\footnotesize
\begin{verbatim}
/* *****  queens-1 ***** */
nDim=8;
bDomain = true;
bNoCapture = true;

for(ni=0; ni<nDim; ni++) {
  bDomain &&= (n[ni]<nDim);
  for(nj=ni+1; nj<nDim; nj++)
    bNoCapture &&= n[ni]!=n[nj] && ni+n[nj]!=nj+n[ni] && ni+n[ni]!=nj+n[nj];
}
assert_all(bDomain && bNoCapture);


*************************************
*********  URSA Interpreter *********
*************************************

--> Solution 1
n[0]=0
n[1]=6
n[2]=3
n[3]=5
n[4]=7
n[5]=1
n[6]=4
n[7]=2

...

--> Solution 92
n[0]=3
n[1]=1
n[2]=4
n[3]=7
n[4]=5
n[5]=0
n[6]=2
n[7]=6

[Formula generation: 0s; conversion to CNF: 0.01s; total: 0.01s]
[Solving time: 0.08s]
[Formula size: 841 variables, 3352 clauses]
\end{verbatim}
}

In the above specification (referred to as to {\tt queens-1} in the
following text), $i$-th row ($i=0,\ldots,N-1$) is associated
with the $i$-th element the array \verb|n|. In each row there should be
one queen and \verb|n[i]| is equal to the column in which that queen is placed.
The variable \verb|bDomain| encodes the condition that each \verb|n[i]| is
between $0$ and $N-1$ and the variable \verb|bNoCapture| encodes the
condition that there are no two queens that attach each other. If the
numbers are represented by vectors of length $l$, for each problem instance
there are $lN$ basic variables in the propositional formula generated.
For $N<16$, the numbers (and intermediate results such as \verb|ni+n[nj]|)
in the above specification can be represented by $5$ bits.
Table \ref{table:queens} shows experimental results for the above specification
for the instance sizes $N=1$, $2$, $3$, $\ldots$, $12$
(instances for which all solutions were found within 600s), including the number
of solutions, the number of variables and clauses in the generated propositional
formula, and the time spent\footnote{
Reported experimental results were obtained on a PC computer with Intel Celeron
420 1.60Ghz, 1Gb RAM, for \ursa using {\sc clasp} as the underlying \sat solver.}
for finding all solutions. The time spent for
finding the first solution was less than $0.05s$ for each problem instance.
The time spent for generating the formulae was less than $0.01s$ for each problem
instance and was negligible compared to the solving time. This shows that the
used generation mechanism is rather efficient.
For these problems (as well as for many other \csp problems), the ratio of
the number of clauses and the number of variables in the generated formulae,
gets rather stable (as the size of the instance increases) and it reflects
the problem {\em constrainedness} \cite{constrainedness}.

\begin{table}[h!]
\begin{center}
{\scriptsize
\begin{tabular}{|l|r|r|r|r|r|r|r|r|r|r|r|r|} \hline
dimension            & 1   & 2   & 3   & 4   & 5    & 6    & 7    & 8    & 9    &  10  & 11    & 12     \\ \hline \hline
number of solutions  & 1   & 0   & 0   & 2   & 10   & 4    & 40   & 92   & 352  &  724 & 2680  & 14200  \\ \hline
number of variables  & 5   & 44  & 115 & 209 & 331  & 480  & 667  & 841  & 1052 & 1286 & 1560  & 1819   \\ \hline
number of clauses    & 5   & 149 & 418 & 794 & 1274 & 1869 & 2612 & 3352 & 4217 & 5179 & 6295  & 7390   \\ \hline
all solutions        & 0   &   0 &   0 &   0 & 0.01 & 0.02 & 0.14 & 0.08 & 0.61 & 3.12 & 19.25 & 116.78 \\ \hline
\end{tabular}
}
\end{center}
\caption{Experimental results for the {\tt queens-1} specification applied for
$N=1, \ldots, 12$}
\label{table:queens}
\end{table}

The above data are obtained using numbers represented by vectors of length $5$.
However, using the minimal vectors length is not critical --- if a larger vector
size is used, only trivial constraints are added and they have a very small impact
on the solving process. Table \ref{table:queens_vectors_length} shows results for the
above specification of the queens problem for instance size $10$, for vectors length
increasing from 5 to 12. The time for generating formulae slightly increases, the
number of variables and clauses increases significantly, but the time spent for
solving remains about the same for all vector lengths.

\begin{table}[h!]
\begin{center}
{\footnotesize
\begin{tabular}{|r|r|r|r|r|r|r|r|r|r|} \hline
number of bits & 5     & 6     & 7     & 8    & 9     & 10    & 11    & 12    \\ \hline \hline
variables      & 1286  & 1611  & 1936  & 2261 & 2586  & 2911  & 3236  & 3561  \\ \hline
clauses        & 5179  & 6604  & 8119  & 9724 & 11419 & 13204 & 15079 & 17044 \\ \hline
generating     & 0.02  & 0.02  & 0.02  & 0.03 & 0.03  & 0.04  & 0.04  & 0.04  \\ \hline
solving        & 3.12  & 3.26  & 3.38  & 3.19 & 3.37  & 3.13  & 3.22  & 3.23  \\ \hline
\end{tabular}
}
\end{center}
\caption{Experimental results for the {\tt queens-1} specification applied (for finding
all solution) for $N=10$ and for number representations using $5,6,\ldots,12$ propositional
formulae}
\label{table:queens_vectors_length}
\end{table}

The same problem can be represented in \ursa in other ways as well. For
instance, the above specification can be slightly modified so it uses
bit-wise operators, instead of arithmetic operators.
Each row of the table can be represented as a $N$-digits number,
i.e., by an element of an array \verb|n|. Each of these numbers should have
exactly one $1$ in its binary representations (which can be checked in the
way described in Example \ref{example:verification}), and at each
position exactly one of \verb|n[i]| should have $1$. The remaining (diagonal)
no-attack conditions (for pairs of positions represented by indexes
(\verb|Ax|,\verb|Ay|) and (\verb|Bx|,\verb|By|)) can be also expressed
by using bit-wise operations. A corresponding \ursa specification
(referred to as to {\tt queens-2} in the following text)
is as follows (it should be used with numbers represented as vectors of length $N$):

\label{page:queens2}
{\footnotesize
\begin{verbatim}
/* *****  queens-2 ***** */
nDim = 8;

bHorizontal = true;
for(ni=0; ni<nDim; ni++)
  bHorizontal &&= ((n[ni] & n[ni]-1)==0) && (n[ni]!=0);

nVertical = 0;
for(ni=0; ni<nDim; ni++)
  nVertical |= n[ni];
bVertical = (nVertical+1 == 0);

bDiagonal = true;
for(nAi=0; nAi<nDim-1; nAi++)
  for(nAj=0; nAj<nDim; nAj++)
    for(nBi=nAi+1; nBi<nDim; nBi++)
      for(nBj=0; nBj<nDim; nBj++)
        if (nBi-nAi==nBj-nAj || nBi-nAi==nAj-nBj)
          bDiagonal &&= (((n[nBj]<<(nBi-nAi)) & n[nAj])==0);

assert_all(bHorizontal && bVertical && bDiagonal);
\end{verbatim}
}

The queens problem can be represented also in the spirit of the {\em direct
encoding}. Namely, each position in the table is associated with one
Boolean variable, so \verb|b[i][j]| is set if and only if the position
$(i,j)$ is occupied by a queen. It has to be ensured that in each row
there is exactly one queen and it has to be ensured that in each column
there is at least one queen (this is sufficient if the former condition
is satisfied). No-attack conditions are expressed in a straightforward
manner. A corresponding \ursa specification is as follows (the numerical
values could be represented by any vector length that can accommodate \verb|nDim|):

{\footnotesize
\begin{verbatim}
/* *****  queens-3 ***** */
nDim = 8;

bHorizontal = true;
for(ni=0; ni<nDim; ni++) {
  bOne = false;
  bMoreThanOne = false;
  for(nj=0; nj<nDim; nj++)  {
    bMoreThanOne ||= bOne && b[ni][nj];
    bOne ||= b[ni][nj];
  }
  bHorizontal &&= bOne && !bMoreThanOne;
}

bVertical = true;
for(ni=0; ni<nDim; ni++) {
  bOne = false;
  for(nj=0; nj<nDim; nj++)
    bOne ||= b[nj][ni];
  bVertical &&= bOne;
}

bDiagonal = true;
for(nAi=0; nAi<nDim-1; nAi++)
  for(nAj=0; nAj<nDim; nAj++)
    for(nBi=nAi+1; nBi<nDim; nBi++)
      for(nBj=0; nBj<nDim; nBj++)
        if (nBi-nAi==nBj-nAj || nBi-nAi==nAj-nBj)
          bDiagonal &&= (!b[nAi][nAj] || !b[nBi][nBj]);

assert_all(bHorizontal && bVertical && bDiagonal);
\end{verbatim}
}

It seems that the four-fold loop in the above specification can be a
source of inefficiency. Indeed, instead of going through all possible values
for \verb|nBi| and only then checking if the corresponding positions should
be tested for attack condition, one can calculate and consider only relevant
coordinates \verb|nBj|, as in the following modified specification:

{\footnotesize
\begin{verbatim}
/* *****  queens-4 ***** */
nDim = 8;

bHorizontal = true;
for(nx=0; nx<nDim; nx++) {
  bOne = false;
  bMoreThanOne = false;
  for(ny=0; ny<nDim; ny++)  {
    bMoreThanOne ||= bOne && b[nx][ny];
    bOne ||= b[nx][ny];
  }
  bHorizontal &&= bOne && !bMoreThanOne;
}

bVertical = true;
for(ny=0; ny<nDim; ny++) {
  bOne = false;
  for(nx=0; nx<nDim; nx++)
    bOne ||= b[nx][ny];
  bVertical &&= bOne;
}

bDiagonal = true;
for(nAx=0; nAx<nDim-1; nAx++)
  for(nBx=nAx+1; nBx<nDim; nBx++)  {
    for(nAy=0; nBx-nAx+nAy<nDim; nAy++)  {
      nBy=nBx-nAx+nAy;
      bDiagonal &&= (!b[nAx][nAy] || !b[nBx][nBy]);
    }
    for(nAy=nBx-nAx; nAy<nDim; nAy++)  {
      nBy=nAy-(nBx-nAx);
      bDiagonal &&= (!b[nAx][nAy] || !b[nBx][nBy]);
    }
  }

assert_all(bHorizontal && bVertical && bDiagonal);
\end{verbatim}
}

However, this modification does not improve efficiency of the solving process.
Namely, all constraints that are generated by the former and not by the later
specification are trivially discarded and the final generated formulae are identical
in the two cases. The formula generation is, still, somewhat more efficient in the
latter case, but that gain gets insignificant as the problem instance grow.

Table \ref{table:queens_specifications} shows experimental results for the
four given representations of the queens problem. For the first representation
numbers are represented by vectors of length $5$, for the second, numbers are
represented by vectors of length equal to the instance size, and for the last
two, numbers are represented by vectors of length $4$. The table shows times
(in seconds) for finding all solutions, while the timeout was set at 600s.
The first problem representation was clearly shown to be the least efficient,
while modified version (the second representation) was the best.
As expected, the third and the fourth specification were almost equal.
According to the data, it appears that the second specification was most
efficient. However, dealing with output of large number of variables
takes a larger portion of time for the last two specifications than
for the second one. If used in a ``quiet'' mode (without listing values of
independent variables), the last two specifications are slightly more
efficient than the second one.
In summary, the specifications based on the bitvector operators
and on the direct encoding were more efficient. Of course, even a conclusion
about the most efficient sort of encoding for the queens problem should
rely on a wider set of specifications and experiments, not to mention
considering most efficient sorts of encodings generally. Although
some encodings typically perform better than some other ones, a full
picture of relative quality of encodings is very complex and each
popular encoding has problems where it performs well. Hence, one of
the points of presenting different specifications of the queens problem
is not to locate the best encodings, but rather to demonstrate that
in the \ursa system one can make significantly different representations
of the same problem.

\begin{table}[h!]
\begin{center}
{\footnotesize
\begin{tabular}{|r|r|r|r|r|r|} \hline
Dimension  & solutions & queens-1 & queens-2 & queens-3 & queens-4  \\ \hline \hline
8          & 92        &  0.08   & 0.08    &  0.09   &   0.03  \\ \hline
9          & 352       &  0.61   & 0.16    &  0.15   &   0.09  \\ \hline
10         & 724       &  3.12   & 0.37    &  0.32   &   0.27  \\ \hline
11         & 2680      & 19.25   & 1.18    &  1.17   &   0.92  \\ \hline
12         & 14200     & 116.78  & 4.92    &  5.52   &   5.15  \\ \hline
13         & 73712     &  --     & 26.47   &  30.85  &  30.48  \\ \hline
14         & 365596    &  --     & 164.19   &  198.06 & 190.45  \\ \hline
\end{tabular}
}
\end{center}
\caption{Experimental results of \ursa applied on the $N$ queens problem for
$N=1, \ldots, 14$ (for finding all solutions) and for four different specifications}
\label{table:queens_specifications}
\end{table}

% ***************************************************************************
% ***************************************************************************
\section{Comparison to Related Techniques, Languages and Tools}
\label{sec:related_work}
% ***************************************************************************
% ***************************************************************************

In this section we discuss tools and techniques related to the presented
approach. We comment on symbolic execution, on constraint solvers, and
on reducing problems to \sat. Finally, we present results of a limited
experimental comparison between \ursa and several other systems.

\subsection{Symbolic execution}

Operation of \ursa is related to symbolic execution. In symbolic execution
\cite{King76,PasareanuV09}, program inputs are represented by symbolic values
rather than by concrete data and the values of program variables are represented
as symbolic expressions. The program is executed by manipulating expressions
involving the symbolic values and, as a result, the output values are expressed
as a function of the input symbolic values. Symbolic execution has been proposed
for software verification over three decades ago, and recently it gained a
renewed interest. The verification tools using symbolic execution include systems like Java
Pathfinder\footnote{\url{http://babelfish.arc.nasa.gov/trac/jpf}} \cite{PasareanuV04},
Pex\footnote{\url{http://research.microsoft.com/en-us/projects/pex/}} \cite{TillmannH08},
Vigilante\footnote{\url{http://research.microsoft.com/en-us/projects/vigilante/}} \cite{CostaCCRZZB08}.
Some of these tools use \sat and \smt solvers, but they typically handle
only machine data-types (and not arbitrary bit-widths). Also, their
purpose is generating test suites and finding (single) models that
lead to bugs (rather than enumerating all solutions of combinatorial
problems).

\subsection{Modelling Languages and Systems}
\label{subsec:languages}

General modelling systems are used for specifying problems in corresponding
modelling languages and solving them by various techniques. There are several
dominating approaches for constraint programming including: constraint logic
programming over finite domains ({\sc clp(fd)}; combines two declarative
programming paradigms -- logic programming and constraint solving) \cite{JaffarM94},
answer set programming ({\sc asp}; a form of declarative programming
with the roots in nonmonotonic reasoning, deductive databases and logic
programming with some {\sc asp} systems using \sat solvers) \cite{ASP-SAT},
and disjunctive logic programming \cite{LeonePFEGPS06}.
There are hybrid systems that use custom specification languages and
provide support for constraint programming (e.g., {\sc ibm ilog opl},
{\sc comet}, {\sc g12}). Also, there are libraries for constraint programming
and combinatorial optimisation for general purpose programming languages:
Ilog solver\footnote{\url{http://ilog.com/products/}} is a C++ library for constraint programming,
Numberjack\footnote{\url{http://4c110.ucc.ie/numberjack/}} is a Python-based
constraint satisfaction and optimisation library with support for
several underlying combinatorial solvers,
$Scala^{Z3}$ is a {\sc Scala} library for checking satisfiability and
solution enumeration with support for the \smt solver Z3 \cite{KoksalKS11}.

Programs in specification languages used by the modelling systems
are generally not directly executed. Rather, they describe a problem at a
high-level, descriptive way and the specification does not say how the problem
is to be solved. In most of modelling languages all solving aspects are
ignored (are stored only in the underlying solver). Modelling systems
typically use custom, different and incompatible modelling languages. There
is no standard modelling language for constraint programming problems.
A language {\sc xcsp} 2.1 is an {\sc xml}-based format to represent various
\csp instances \cite{abs-0902-2362}. The main objective of this language is
to ease the effort required to test and compare different algorithms by
providing a common test-bed of constraint satisfaction instances. {\sc xcsp}
is already used in \csp competitions as an standard input
format.\footnote{\url{http://cpai.ucc.ie/09/}}
A high-lever language {\sc MiniZinc} (a subset of a language {\sc Zinc}
\cite{MarriottNRSBW08}) also aims at becoming a standard specification language
\cite{NethercoteSBBDT07}. {\sc MiniZinc} models can be translated to
{\sc FlatZinc}, a low-level solver input language. There are a number of
differences between the languages {\sc xcsp}, {\sc MiniZinc} and \ursa.
{\sc xcsp} representations are low-level, while {\sc MiniZinc} and \ursa
representations are high-level --- the former is rather a machine-oriented,
interchange format, while the latter two provide high-level, human-readable
specifications. In contrast to {\sc MiniZinc} and \ursa, {\sc xcsp} has
no arrays and looping constructs and in {\sc xcsp}, for each instance,
domains, variables, relations, predicates and constraints are exhaustively
listed.

All specification languages used in the above systems
are based on some form of declarative/logic programming, while \ursa uses
a novel combination of declarative and imperative paradigms.
For some problems, \ursa specifications may be longer than of other systems,
but some problems naturally expressed in \ursa (for instance, problems that
involve bit-wise operations, arithmetic modulo $2^n$, or software verification
problems that involve destructive assignments) are very difficult
(or practically impossible for large scale instances) to express in other
specification languages. Loop constructs are naturally expressed in \ursa,
while their absence in some declarative languages (some specification languages
do admit complex looping structures) may cause a range of difficulties
\cite{LogicalLoops}.\footnote{There are recent ideas for introducing loops
in constraint programming in an imperative language style in order to enable
prototyping new constraints with less effort \cite{LoopsInCP}.}
The \ursa language is expressive enough to enable
substantially different encodings of the same problem, which is often not easy
with other systems. Learning the \ursa language should be trivial to someone
familiar with some widely used imperative programming language such as C or
Java, as there are no specific commands or flow-controls aimed at constraint
solving.
Concerning the solving mechanism, \ursa, like other systems, builds up
a formula representing a constraint and then posts the constraint to an underlying
solver. \ursa's specific is that building this formula is completely specified
by \ursa's simple semantics, corresponding to binary representation of numbers.
Concerning underlying solvers, \ursa uses \sat solvers and differs from the
above described tools that use mathematical programming, constraint logic
programming, and other techniques (with exception of {\sc asp} systems
such as {\sc Cmodels} that also uses \sat solvers).

Example \ref{example:verification} illustrates one family of problems that
can be simply solved by the \ursa system. Specifications of functions
in C (analyzed if they are equivalent) can be almost in verbatim used
within \ursa, while it would be extremely difficult to represent this kind
of problems in existing declarative programming languages --- consider, for
instance, complex functions (say, like cryptographic functions) that
involve a large number of destructive assignments in their specifications.
Additional problem is that the bitwise operators are not supported in
other modelling systems.

\begin{exa}
\label{example:verification}
One of the common problems in software verification is ensuring that two
implementations are equivalent. The \ursa system can be suitable for such
tasks. Consider, for example, two implementations (both based on
C-style bitwise operators) of the check that the input number \verb|n|
has in its representation exactly one bit set: the first
(encoded by {\tt b1}) is rather compact, while the second
(encoded by {\tt b2}) is more elaborated but simpler.
\ursa (for the vector length for representation of numbers equal \verb|nLen|),
verifies that the condition \verb|b1^^b2| is unsatisfiable, hence the two
implementations give the same result for any input number.

{\footnotesize
\begin{verbatim}
b1 = (nv!=0) && ((nv & (nv-1))==0);

nLen=8;
bOne = false;
bMoreThanOne = false;
for(ni=0; ni<nLen; ni++)  {
  bMoreThanOne ||= bOne && ((nv & 1)!=0);
  bOne ||= ((nv & 1)!=0);
  nv >>= 1;
}
b2 = bOne && !bMoreThanOne;

assert(b1^^b2);
\end{verbatim}
}
\end{exa}

\subsection{Reduction to \sat and \sat-Based Constraint Solvers}
\label{subsec:sat_based_solvers}

There are a huge number of problems solved by reduction to \sat, in a
range of domains (e.g., in scheduling \cite{scheduling-SAT}, termination
analysis \cite{termination-SAT}, cryptanalysis \cite{Massacci00,MironovZ06},
model checking \cite{ClarkeBRZ01}, to name just a few).
One of the reasons for such a wide application range are tremendous
advances in \sat technology over the last years and a number of efficient
solvers. The \ursa system does not introduce or promote one \sat solver.
Rather, \ursa can use any \sat solver in the solving phase. Moreover, it
is fruitful to have, within \ursa, a number of different \sat solvers,
appropriate for different sorts of input problems, following ideas of
\sat portfolio solvers \cite{NikolicMJ09}.

There are several approaches for encoding \csp problems into \sat
\cite{Prestwich09}. Probably the most popular basic types of encodings
into \sat are: the {\em sparse encoding}, the {\em compact encoding}, and
the {\em order encoding} \cite{Hoos99,TamuraTKB09}. In the sparse
encoding, a propositional variable $x_{v,i}$ is defined as true iff
the integer variable $v$ has a value $i$ assigned to it. Examples
of the sparse encoding are the {\em direct encoding} and the
{\em support encoding} \cite{Prestwich09}. In the compact encoding
(or {\em log encoding}), a propositional variable is assigned to each
bit of each integer variable (within a finite domain) \cite{Prestwich09}.
In the order encoding (also known as {\em regular encoding}), related
to many-valued logics and often used for the finite-domain linear problems,
an inequality $v \leq i$ is encoded by a different propositional variable
for each integer variable $v$ and integer value $i$ \cite{TamuraTKB09,ArgelichCLM09}.
Even within one encoding style, modelling of a problem can take significantly
different forms \cite{SmithSW00}. There are a number of both theoretical
and practical studies and comparisons between different encoding schemes.
Since the log encoding lacks the propagation power of the direct and
support encodings, it typically leads to less efficient solutions,
compared to these two \cite{Walsh00,Gelder08}.
The order encoding gives better performance compared with the direct
encoding and the support encoding for some \csp problems \cite{TamuraTKB09}.
Within the \ursa system, numerical variables are represented using
binary representation that corresponds to the compact encoding, but
still, thanks to the expressiveness of the specification language,
other encoding styles can be simulated and used (as shown by the presented
examples).

The \ursa system is related to the special-purpose system for transforming
cryptanalysis of hash functions into the \sat problem \cite{hash-functions}.
In that approach, implementations
of hash functions in C++ were used and, by overloading the standard arithmetic
and logical operators in C++ and by running the code of the hash functions
within such framework, propositional formulae corresponding to cryptanalysis
tasks were generated (and then solved by a \sat solver). Similar approach was
also used for cryptanalysis of \des \cite{milan_sesum_msc}. \ursa is a general
framework aimed not only to cryptanalysis tasks, but to a much wider range of
problems. The framework consists of both a modelling language (instead of C++)
and the solving machinery, tightly integrated. The system is stand-alone, does
not involve the C++ language in the modelling process, and the language itself
defines the modelling power of the approach. The related, C++ based
approach, has an advantage that the specification can be used directly as a
C++ code, but \ursa specifications can often be used as C code in verbatim.

There are several general-purpose constraint solving systems based on \sat.
Thanks to the advances in \sat technology, such systems can be very efficient,
despite the main weakness of this approach: a domain knowledge and a global
structure of the problem are lost when it is reduced to the simple propositional
logic.
{\sc sugar}\footnote{\url{http://bach.istc.kobe-u.ac.jp/sugar/}} is a \sat-based
constraint solver that uses the order encoding \cite{TamuraTKB09}. It is
focused on compiling finite linear \csp into \sat. {\sc sugar} uses a syntax
of \csp that is designed to cover the notation of the {\sc xcsp} 2.1 format.
{\sc FznTini}\footnote{\url{http://users.cecs.anu.edu.au/~jinbo/fzntini/}}
is a general constraint solver that solves constraint satisfaction and
optimization problems (not involving floating point numbers) given in the
general constraint language {\sc FlatZinc} (typically produced from {\sc MiniZinc}
specifications) by translating them to \sat and calling a \sat solver {\sc Tinisat}
\cite{Huang08}. The system can be also used for translating to \sat only (so,
it can be used by independent \sat solvers). {\sc FznTini} uses a fixed encoding
directly related to binary representation of integers (the two's complement
representation) that corresponds to the compact encoding.
{\sc NPSpec} is a modelling language for constraint problems, going with a tool
{\sc Spec2Sat} that compiles specifications into \sat instances \cite{NPSpec}.
{\sc NPSpec} uses a highly declarative style of programming, similar to Datalog
(a query language for deductive databases). The semantics of {\sc NPSpec} is
based on the model minimality, an extension of the least fixed point semantics
of the Horn fragment of first order logic \cite{NPSpec}.
The system {\sc mxg}\footnote{\url{http://www.cs.sfu.ca/research/groups/mxp/}},
focused on \np-hard problems, uses a modelling language based on classical first
order logic, and for a given specification produces a propositional formula (and
passes it to a \sat solver) \cite{MitchellT05,PelovT05}. The system can also
translate problem specification to \sat extended with cardinality constraints.

Apart from systems that can translate problem specifications to \sat, there
are also systems that translate such specifications to \smt. For instance,
{\sc fzn2smt} \cite{fzn2smt} translates from the {\sc FlatZinc} language
and the system {\sc simply} \cite{SIMPLY_CP_SMT} translates from a
declarative modelling language similar to, but simpler than {\sc MiniZinc}.
Both systems translate to the standard {\sc smt-lib} format\footnote{\url{http://www.smtlib.org/}}
and can use various underlying \smt theories and available \smt solvers.

The system \ursa and the above systems share the underlying solving technology,
but their specification languages are very different. In contrast to \ursa, all
of the above tools use declarative languages, and they don't have features of
imperative languages (e.g., destructive assignments), as discussed in Section
\ref{subsec:languages}. Also, the languages used by the above tools provide
support for various operators and global constraints, but typically do not
support bit-wise operators and constraints involving modular arithmetic,
which can be essential in many cases. This way, the input languages restrict
the tools from having the full power of modelling in propositional logic.
Concerning expressiveness, {\sc sugar} and {\sc FznTini} use rather simple,
low-level specification language without flow control structures and other
features of programming languages that \ursa has. Using representation based
on binary representation of integers is very similar in {\sc FznTini} and \ursa
(with a minor difference that {\sc FznTini} uses the two's complement
representation).

\subsection{Experimental Comparison}
\label{subsec:experimental}

It is very difficult to make a fair and thorough comparison of the modelling
systems: they were not built with the same motivation and purposes; it is not
only performance that is important but also expressiveness and ease of acquiring
a modelling language; some systems perform better on some sorts of problems
(and worse on the other); a single problem can be often specified in different
ways; most systems do not share the same input language (even if they do, some
types of specifications may be better suited to some systems); the leading
systems are under ongoing developments, and there are new features being added
constantly; there are new emerging systems, etc. Still, with all of the above
cautious, several (a very few, likely due to the above difficulties) existing
reports give some general picture of efficiency of systems for constraint
solving over finite domains \cite{FernandezH00,ManciniMPC08,DovierFP09} while
new insights are provided by a number of system competitions recently initiated.

For experimental comparison with the \ursa system, we used the following
state-of-the-art systems (and versions), including several industrial ones:
constraint logic programming systems
{\sc B-Prolog} 7.4\footnote{\url{http://www.probp.com/}} and
{\sc SICStus} 4.1.1\footnote{\url{http://www.sics.se/isl/sicstuswww/site/}} (with the \verb|clpdf| module),
a deductive database system (using Disjunctive Datalog language that
combines databases and logic programming) {\sc dlv}\footnote{\url{http://www.dbai.tuwien.ac.at/proj/dlv}}
\cite{LeonePFEGPS06},
{\sc asp} solvers (used with {\sc Smodels} specifications and a grounder
{\sc Lparse} 1.1.2)
{\sc clasp} 1.3.2\footnote{\url{http://www.cs.uni-potsdam.de/clasp/}}
\cite{GebserKNS07a},
{\sc Cmodels} 3.79\footnote{\url{http://www.cs.utexas.edu/~tag/cmodels/}}
\cite{GiunchigliaLM06} and
{\sc Smodels} 2.34\footnote{\url{http://www.tcs.hut.fi/Software/smodels/}}
\cite{SyrjanenN01},
hybrid optimization system (using a custom object-oriented programming language)
{\sc comet} 2.1.1\footnote{\url{http://dynadec.com}} \cite{MichelH05a},
a hybrid optimization system (using a custom declarative modelling language)
{\sc ibm ilog opl 6.3}\footnote{\url{http://www-01.ibm.com/software}},
a default finite-domain
{\sc g12/fd} solver Mercury (using {\sc MiniZinc} as
its input language) from MiniZinc 1.1.1\footnote{\url{www.g12.cs.mu.oz.au/minizinc/}}
\cite{StuckeyBMMSSWW05},
and \sat-based systems
{\sc sugar} 1.14.6
and {\sc FznTini}.\footnote{We didn't include the systems {\sc mxg} and
{\sc NPSpec}/{\sc Spec2Sat} in this evaluation: {\sc mxg} was not publicly
available and {\sc NPSpec}/{\sc Spec2Sat} was not maintained since 2005, and also
its reported performance \cite{NPSpec2006}, especially for the \sat formulae
generation phase, is significantly poorer than of \ursa and other considered systems.}
The \ursa system was used with {\sc clasp} 1.2.0 as an underlying \sat solver.

We performed experimental comparison between all of the above tools on a
prototypical \csp problem --- the queens problem (that involves different
sorts of constraints), and additional comparison between the systems that
performed the best in the first phase.
Most of the used specifications are given as part of the system distributions;
they are typically straightforward for their systems and symmetry breaking
conditions or some other additional constraints were not used. Experiments
were performed on a PC computer with Intel Celeron 420 1.60Ghz, 1Gb RAM
and the time threshold for finding all solutions was 600s.
The measured times are reported by the systems themselves or by adding the
spent ``user'' and ``system'' time.\footnote{For some recommendations on how
to benchmark constraint solving systems visit
\url{http://www.dbai.tuwien.ac.at/proj/dlv/bench/}.}

\paragraph{The Queens problem.}
Table \ref{table:comparison} shows results of experiments for the queens
problem for instance sizes from 8 to 14. The \ursa system was used with
the specification \verb|queens-2| (see page \pageref{page:queens2}).
The most efficient system on this benchmark was {\sc B-Prolog}, followed
by {\sc SICStus}, and then by \ursa, that had similar performance as
{\sc g12/fd} and {\sc clasp}. It is interesting to notice that the
\ursa system using {\sc clasp} as an underlying \sat solver was around the
same efficiency as the {\sc clasp} solver used as {\sc asp} solver, which
suggests that the reduction to \sat used by \ursa is very efficient.

\begin{table}[h!]
\begin{center}
{\scriptsize
\begin{tabular}{|r|r|r|r|r|r|r|r|r|r|r|r|} \hline
$N$ & solutions & {\tiny\sc Cmodels} & {\tiny\sc Smodels} & {\tiny\sc dlv} & {\tiny\sc opl} & {\tiny\sc comet} & {\tiny\sc clasp}& {\tiny\sc g12/fd} & {\tiny \ursa}   & {\tiny\sc SICStus} & {\tiny\sc B-prolog} \\ \hline \hline
8   & 92        & 0.11          &  0.09         &  0.04     &   0.08 &   0.02  &   0.02     &   0.03       &   0.08 &    0.01        &   0.01   \\ \hline
9   & 352       & 0.75          & 0.35          &  0.15     &   0.30 &   0.09  &   0.07     &   0.08       &   0.16 &    0.03        &   0.01   \\ \hline
10  & 724       & 7.42          & 1.41          &  0.45     &   0.63 &   0.34  &   0.22     &   0.22       &   0.37 &    0.09        &   0.02   \\ \hline
11  & 2680      & 132.20        & 10.51         &  2.31     &   2.28 &   1.64  &   0.89     &   1.12       &   1.18 &    0.50        &   0.10   \\ \hline
12  & 14200     & $>$600        & 44.37         & 12.67     &  12.56 &   9.34  &   4.51     &   5.91       &   4.92 &    2.53        &   0.45   \\ \hline
13  & 73712     & $>$600        & 331.54        & 83.36     &  62.89 &  48.73  &  25.50     &  31.48       &  26.47 &   14.36        &   2.85   \\ \hline
14  & 365596    & $>$600        & $>$600        &479.91     & 301.88 & 246.90  & 190.48     & 188.43       & 164.19 &   76.89        &  17.64   \\
\hline
\end{tabular}
}
\end{center}
\caption{Results of experimental comparison of ten tools (including \ursa)
applied on the $N$ queens problem for $N=8,\ldots, 14$}
\label{table:comparison}
\end{table}

The above result don't include the systems that translate problem specifications
to \sat and which are the systems closest to \ursa. Namely, these systems translate
inputs to \sat (so it can be considered that they share the solving mechanism),
but they use different \sat solvers. A fair comparison would be thus to use these
systems only as translators to \sat and then use the same \sat solver (for instance,
{\sc clasp}) for finding all models of the generated \sat formulae. It is interesting
to consider size of generated formulae and solving times (of course, smaller formulae
does not necessarily lead to shorter solving times). {\sc FznTini} was used with {\sc FlatZinc}
specifications obtained from the {\sc MiniZinc} specification used by {\sc g12/fd}
(with integers encoded with 5 bits) and with {\sc FlatZinc} specifications obtained
from a {\sc MiniZinc} specification made in the style of the direct
encoding,\footnote{Therefore, these translations to \sat are rather by
two systems: the {\sc MiniZinc} to {\sc FlatZinc} converter and {\sc FznTini}.}
we will denote them by 1 and 2. {\sc sugar} was used only with a specification
that employs the order encoding. \ursa was used with the
specifications \verb|queens-1| (with integers encoded with 5 bits),
\verb|queens-2| (with the number of bits equal the instance dimension),
and \verb|queens-3|, (with integers encoded with 4 bits).
Table \ref{table:sugar_ursa} presents the obtained experimental results.
All recorded times were obtained for the ``quiet'' mode of the \sat solver
(without printing the models). Times for generating formulae were negligible
(compared to the solving phase) for all systems, so we don't report them here.

For related specifications, \ursa's \verb|queens-1| gave much smaller formulae
(probably thanks to techniques mentioned in Section \ref{subsec:translation_to_sat})
and somewhat better performance than {\sc FznTini} 1, which suggests that
{\sc FznTini} does not benefit much from information about the global structure
of the problem. The formulae generated by {\sc sugar} were significantly smaller
than in the above two cases, and led to much better solving efficiency. However,
it was outperformed by the remaining entrants. The \ursa's specifications \verb|queens-2|
and \verb|queens-3| gave similar results. The specification \verb|queens-3| produced
formulae with the smallest number of clauses. {\sc FznTini} 2 produced formulae
with the smallest number of variables. The best results in terms of the solving
times were obtained also by {\sc FznTini} 2. It can be concluded that \ursa can
produce, with suitable problem specifications, propositional formulae comparable
in size and in solving times with formulae produced by related state-of-the-art systems.

\begin{table}[h!]
\begin{center}
{\footnotesize
\begin{tabular}{|r|r|r|r|r|r|r|r|} \hline
Dimension     &    8 &    9  &   10  &   11  &   12   &    13  & 14      \\ \hline
\multicolumn{8}{l}{{\sc FznTini} 1:}                                     \\ \hline
variables     & 3012 & 3825  & 4735  & 5742  & 6846   & 8047   & 9345    \\ \hline
clauses       & 9128 & 11628 & 14460 & 17567 & 21000  & 24713  & 28770   \\ \hline
all solutions & 0.15 & 0.79  & 3.20  & 14.53 & 111.78 & $>$600 & $>$600  \\ \hline
\multicolumn{8}{l}{\ursa (queens-1):}                                    \\ \hline
variables     & 841  & 1052  & 1286  & 1560  & 1819   &  2139  & 2468    \\ \hline
clauses       & 3352 & 4217  & 5179  & 6295  & 7390   &  8712  & 10089   \\ \hline
all solutions & 0.08 & 0.53  & 2.83  & 17.60 & 98.04  & $>$600 & $>$600  \\ \hline
\multicolumn{8}{l}{{\sc sugar}:}                                         \\ \hline
variables     & 220  &  284  &  356  &  436  &  524   &  620   &  724    \\ \hline
clauses       & 1138 & 1653  & 2253  & 3012  & 3924   & 5003   & 6263    \\ \hline
all solutions & 0.02 & 0.06  & 0.31  & 1.58  & 9.59   & 68.07  & 411.15  \\ \hline
\multicolumn{8}{l}{\ursa (queens-2):}                                    \\ \hline
variables     & 542  &  739  & 978   & 1263  & 1598   &  1987  & 2434    \\ \hline
clauses       & 3319 & 5008  & 7280  & 10258 & 14077  & 18884  & 24838   \\ \hline
all solutions & 0.01 & 0.04  & 0.12  & 0.70  & 4.01   & 23.17  & 138.55  \\ \hline
\multicolumn{8}{l}{\ursa (queens-3):}                                    \\ \hline
variables     & 176  &  225  & 280   &  341  & 408    &  481   & 560     \\ \hline
clauses       & 800  & 1110  & 1490  & 1947  & 2488   &  3120  & 3850    \\ \hline
all solutions & 0.01 & 0.03  & 0.12  & 0.69  & 3.82   & 21.09  & 136.45  \\ \hline
\multicolumn{8}{l}{{\sc FznTini} 2:}                                     \\ \hline
variables     & 128  & 162   & 200   & 242   & 288    &   338  & 392     \\ \hline
clauses       & 872  & 1236  & 1690  & 2244  & 2908   &  3692  & 4606    \\ \hline
all solutions & 0.01 & 0.02  & 0.07  & 0.33  & 1.52   &  8.39  & 52.12   \\ \hline
\end{tabular}
}
\end{center}
\caption{Data on \sat formulae generated by {\sc FznTini}, {\sc sugar} and \ursa for
the $N$ queens problem for $N=8,\ldots, 14$ and solved by the {\sc clasp} \sat solver}
\label{table:sugar_ursa}
\end{table}

\paragraph{Additional Experiments.}
In additional experiments, only the systems that performed the best on the
queens problem were considered (with only one constraint logic programming
system kept): {\sc B-prolog}, {\sc clasp}, {\sc FznTini}, {\sc g12/fd}, and \ursa.
The following problems were considered (for all problems all solutions were
sought):

\begin{desCription}
\item\noindent{\hskip-12 pt\bf Golomb Ruler:}\
The problem (actually, one of its variation) is as follows, given a value $L$
check if there are $m$ numbers $a_0$, $a_1$, $\ldots$, $a_{m-1}$ such that
$0 = a_0 < a_1 < ... < a_{m-1} = L$ and all the $m(m-1)/2$ differences
$a_j - a_i$, $0 \leq i < j \leq m$ are distinct (problem 6 at
CSPlib\footnote{\url{http://www.csplib.org/}}).
The experiments were performed for $m=5,\ldots,11$, with the largest $L$
that make the problem unsatisfiable and with the smallest $L$ that make
the problem satisfiable.

\item\noindent{\hskip-12 pt\bf Magic Square:}\
A magic square of order $N$ is a $N \times N$ matrix containing the numbers
from $0$ to $N^2-1$, with each row, column and main diagonal equal the same sum.
The problem is to find all magic squares of order $N$ (problem 19 at
CSPlib).
The experiments were performed for $N=3$ and $N=4$.

\item\noindent{\hskip-12 pt\bf Linear Recurrence Relations:}\
Linear homogeneous recurrence relations of degree $k$ are of the form:
$T_{n+k+1}=a_{k}T_{n+k}+\ldots+a_1 T_{n+1}$  for $n \leq 0$. Given $T_1$,
\ldots, $T_{k}$ and $n$, $T_n$ can be simply calculated, but finding explicit
formula for $T_i$ requires solving a nonlinear characteristic equation of degree
$k$, which is not always possible. So, the following problem is nontrivial:
given $T_1$, $\ldots$, $T_{k-1}$ and $T_n$, find $T_k$. For the experiment,
we used the relation $T_{n+3} = T_{n+2} + T_{n+1} + T_n$, $T_1=T_2=1$. We generated
instances with the (only) solution $T_3=1$ and the systems were required to seek
all possible values for $T_3$. Additional constraints (used explicitly or
implicitly) for all considered systems) were $T_i > 0$ and $T_i \leq T_n$,
for $i=1,\ldots,n$. \ursa and {\sc FznTini} were used with 32 bit length
for numerical values.

\item\noindent{\hskip-12 pt\bf Non-linear recurrence relations:}\
In non-linear homogenous recurrence relations
of degree k, the link between $T_{n+k}$ and $T_{n+k-1}$, $T_{n+k-2}$, $\ldots$, $T_n$,
is not necessarily linear. For the experiment we used the relation
$T_{n+4} = T_{n+3} \cdot T_{n+2} + T_{n+1}$, $T_1=T_2=1$. We generated
instances with the (only) solution $T_3=1$ and the systems were required to seek
all possible values for $T_3$. Additional constraints (used explicitly of
implicitly) for all considered systems) were $T_i>0$ and $T_i \leq T_n$,
for $i=1,\ldots,n$. For all problem instances, the size of all relevant values
$T_i$ were smaller than $2^{32}$. \ursa and {\sc FznTini} were used with 32 bit length
for numerical values.
\end{desCription}

\noindent The \ursa was used with the following specification for the
Golomb ruler problem (for the instance $m=7$, $L=25$):

{\footnotesize
\begin{verbatim}
nM=7;
nL=25;
bRulerEndpoints = num2bool(nRuler & nRuler >> nL & 1);

nMarks=2;
bDistanceDiff=true;
for(ni=1; ni<=nL-1; ni++) {
  nMarks += (nRuler >> ni) & 1;
  n = (nRuler & (nRuler << ni));
  bDistanceDiff &&= (n & (n-1))==0;
}

assert_all(bRulerEndpoints && nMarks==nM && bDistanceDiff);
\end{verbatim}
}

The above specification employs a binary representation of the ruler (\verb|nRuler|) where each
bit set denotes a mark. The value \verb|nRuler & nRuler >> nL & 1| equals \verb|1| if and only
if the first and \verb|nL|th bit are set (as the ruler endpoints). The value \verb|nMarks| counts
the bits set (e.g., the marks) and it should equal \verb|nM|.
If the ruler \verb|nRuler| is a Golomb ruler, then whenever it is shifted left (for values
\verb|1|, $\ldots$, \verb|nL-1|) and bitwise conjunction is performed with the original ruler
giving the value \verb|n|, there will be at most one bit set in \verb|n| (since all the
differences between the marks are distinct). There is at most one bit set in \verb|n| if
and only if the value \verb|n & (n-1)| equals 0. This specification, employing a single
loop, illustrate the expressive power of bitwise operations supported in the \ursa language.

For the magic square problem, \ursa was used with the following specification:

{\footnotesize
\begin{verbatim}
nDim=4;
nN=nDim*nDim;
bCorrectSum = (2*nSum*nDim == nN*(nN-1));

bDomain=true;
bDistinct=true;
for(ni=0;ni<nDim;ni++) {
  for(nj=0;nj<nDim;nj++) {
    bDomain &&= (nT[ni][nj]<nN);
    for(nk=0;nk<nDim;nk++)
      for(nl=0;nl<nDim;nl++)
        bDistinct &&= ((ni==nk && nj==nl) || nT[ni][nj]!=nT[nk][nl]);
  }
}

bSum=true;
nSum1=0;
nSum2=0;
for(ni=0;ni<nDim;ni++) {
  nSum1 += nT[ni][ni];
  nSum2 += nT[ni][nDim-ni-1];
  nSum3=0;
  nSum4=0;
  for(nj=0;nj<nDim;nj++) {
    nSum3 += nT[ni][nj];
    nSum4 += nT[nj][ni];
  }
  bSum &&= (nSum3==nSum);
  bSum &&= (nSum4==nSum);
}
bSum &&= (nSum1==nSum);
bSum &&= (nSum2==nSum);

assert_all(bCorrectSum && bDomain && bDistinct && bSum);
\end{verbatim}
}

For the linear recursive relation, \ursa was used with the following specification
(the specification for the non-recursive relations is analogous):

{\footnotesize
\begin{verbatim}
n  = 30;
ny = 20603361;

n1=1;
n2=1;
n3=nx;

for(ni=4; ni <= n; ni++) {
  ntmp=n1+n2+n3;
  n1=n2;
  n2=n3;
  n3=ntmp;
  bDomain &&= (n3<=ny);
}

assert_all(bDomain && n3==ny);
\end{verbatim}
}

Table \ref{table:additional_experiments} shows experimental results
(with translation times included).
{\sc FznTini} was used with {\sc clasp} as an underlying solver (the built-in
solver gave poorer results). The number of variables and clauses generated
by \ursa were, in these benchmarks, smaller than for {\sc FznTini}.
For {\sc lparse/clasp}, the translation time was significant, and some of the poor results
for some benchmarks are due to large domains (while the system works with
relations rather than functions). For recurrence relations, {\sc g12/fd}
reported model inconsistency when it approached its limit for integers,
while {\sc B-prolog} just failed to find a solution for larger instances.
Overall, on this set of benchmarks, \ursa gave better results than
{\sc clasp} and {\sc FznTini} and on some benchmarks outperformed all
other tools.

%Size of formulae for Golomb ruler:
%Dimension  & 5/10       & 6/16        & 7/24        & 8/33        & 9/43       & 10/54       & 11/71          \\ \hline \hline
%URSA       & 312/1525   & 792/4666    & 1768/12774  & 3325/29010  & 5625/58450 & 8848/108159 & 15257/229726  \\ \hline
%fzntini    & 3988/10592 & 7494/19901  & 12931/34433 & 21223/56326 &           &            &               \\ \hline

%Size of formulae:
%Dimension  & 5/11(4)    & 6/17(8)     & 7/25(10)    & 8/34(2)     & 9/44(2)    & 10/55(2)    & 11/72(4)
%URSA       & 377/1906   & 893/5410    & 1917/14170  & 3528/31369  & 5888/62164 & 9177/113710 & 15688/238830 \\ \hline
%fzntini    & 3988/10586 & 7494/19891  & 12931/34418 & 21223/56396 &           &             &              \\ \hline

\begin{table}[h!]
\begin{center}
{\footnotesize
\begin{tabular}{|r||r|r|r|r|r|} \hline
Dimension  & {\sc B-prolog}  & {\sc clasp}  & {\sc FznTini} & {\sc g12/fd} & \ursa     \\ \hline \hline
\multicolumn{6}{|c|}{Golomb ruler}                            \\ \hline \hline
5/10       & 0.01     &  5.36  & 0.20    & 0.06   &  0.01     \\ \hline
6/16       & 0.01     & 44.68  & 1.16    & 0.08   &  0.02     \\ \hline
7/24       & 0.01     & 350.11 & 9.53    & 0.10   &  0.10     \\ \hline
8/33       & 0.08     & $>$600 & 111.90  & 0.24   &  0.69     \\ \hline
9/43       & 0.69     & $>$600 & $>$600  & 1.18   &  4.89     \\ \hline
10/54      & 5.34     & $>$600 & $>$600  & 7.84   &  35.55    \\ \hline
11/71      & 105.54   & $>$600 & $>$600  & 93.2   &  571.90   \\ \hline \hline
5/11       & 0.01     &  6.33  &  0.32   & 0.07   &  0.01     \\ \hline
6/17       & 0.01     & 57.40  &  1.43   & 0.08   &  0.01     \\ \hline
7/25       & 0.01     & 429.98 & 14.15   & 0.10   &  0.13     \\ \hline
8/34       & 0.09     & $>$600 & 106.26  & 0.26   &  0.87     \\ \hline
9/44       & 0.78     & $>$600 & $>$600  & 1.27   &  5.87     \\ \hline
10/55      & 6.86     & $>$600 & $>$600  & 6.44   &  37.28    \\ \hline
11/72      & 115.81   & $>$600 & $>$600  & 125.50 &  450.41   \\ \hline \hline
\multicolumn{6}{|c|}{Magic square}                            \\ \hline \hline
3          & 0.01     & 0.05   & 0.04    & 0.01   &  0.01     \\ \hline
4          & 4.74     & 462.21 & $>$600  & 10.26  &  93.01    \\ \hline \hline
\multicolumn{6}{|c|}{Linear recurrence relations}            \\ \hline \hline
4          &     0.00 &  0.01  &   0.01  &  0.01  &  0.00    \\ \hline
5          &     0.00 &  0.06  &   0.01  &  0.02  &  0.00    \\ \hline
6          &     0.00 &  1.18  &   0.01  &  0.02  &  0.01    \\ \hline
7          &     0.00 & 25.49  &   0.01  &  0.02  &  0.01    \\ \hline
$\ldots$   &          &        &         &        &          \\ \hline
28         &    43.83 & $>$600 &   0.17  & 143.54 &  0.31    \\ \hline
29         &    84.92 & $>$600 &   0.68  & incons &  0.33    \\ \hline
30         &   158.78 & $>$600 &   1.02  & incons &  0.47    \\ \hline \hline
\multicolumn{6}{|c|}{Non-linear recurrence relations}        \\ \hline \hline
4          &     0.00 &  0.00  & 0.01    &  0.01  &  0.00      \\ \hline
5          &     0.00 &  0.01  & 0.01    &  0.01  &  0.02    \\  \hline
6          &     0.00 &  0.42  & 0.22    &  0.02  &  0.05    \\ \hline
7          &     0.00 &  126.05& 0.36    &  0.02  &  0.08    \\ \hline
8          &     0.00 & $>$600 & 0.51    &  0.02  &  0.13    \\ \hline
9          &     fail & $>$600 & 0.76    &  0.02  &  0.28    \\ \hline
10         &     fail & $>$600 & 0.88    &  0.03  &  0.53    \\ \hline
11         &     fail & $>$600 & 0.97    &  incons&  0.77    \\  \hline
\end{tabular}
}
\end{center}
\caption{Results of experimental comparison of five tools}
\label{table:additional_experiments}
\end{table}

\paragraph{Discussion.}
The described limited experiments cannot give definite conclusions or ranking
of the considered systems, as discussed above. In particular, one may raise
the following concerns, that can be confronted with the following arguments:

\begin{iteMize}{$\bullet$}
\item {\em \ursa was used with a good problem specifications, and there
may be specifications for other systems that lead to better efficiency.}
However, almost all specifications were taken from the system
distributions, given there to illustrate the modelling and solving
power of the systems. Also, the problem specifications used for \ursa
are also probably not the best possible, but are rather straightforward,
as specifications for other systems. In addition, in contrast to \ursa,
other modelling systems typically aim at liberating the user of thinking
of details of internal representations and are free to perform
internal reformulations of the input problem. Making specifications in
\ursa may be somewhat more demanding than for some other systems, but
gives to the user a fuller control of problem representation.

\item {\em Some specifications used for \ursa are related to the direct
encoding (known to be efficient, for example, for the queens problem),
while this is not the case with other systems.} What is suitable for
\sat-based systems is not necessarily suitable for other systems. For
instance, {\sc g12/fd} gives significantly poorer results with the
specification of the queens problem based on the direct encoding,
than with the one used in the experiment. This is not surprising, because
systems that are not based on \sat does not necessarily handle efficiently
large numbers of Boolean variables and constraints (in contrast to \sat
solvers) and lessons from the \sat world (e.g., that for some sort of
problems, some encoding scheme is the most efficient) cannot be {\em a priori}
applied to other solving paradigms.

\item {\em \ursa uses bit-wise logical operators, while other systems
do not (as they don't have support for them).} Bit-wise logical operators
make one of advantages of \ursa, while in the same time, some other
systems use their good weapons (e.g., global constraints such as
all-different).
\end{iteMize}

In summary, the presented experiments suggest that
the \ursa system (although it is not primarily a \csp solver but a general system
for reducing problems to \sat) is competitive to the state-of-the-art,
both academic and industrial, modelling systems --- even if they can encode
high-level structural information about the input problem and even
if they involve specialized underlying solvers (such as support for
global constraints like all-different).
A wider and deeper comparison between these (and some other) constraint
solvers (not sharing input language) and with different encodings of
considered problems, would give a better overall picture but is out
of scope of this paper.

% ***************************************************************************
% ***************************************************************************
\section{Future Work}
\label{sec:future_work}
% ***************************************************************************
% ***************************************************************************

The current system (with the presented semantics and the corresponding
implementation) uses one way (binary representation) for representing (unsigned)
integers but (as shown in the given examples), it still enables using different
encoding styles in specifications. For further convenience, we are planning to
natively support other representations for integers, so the user could choose
among several encodings. Also, signed integers and floating point numbers could
be supported.

The language \ursa (and the interpreter) can be extended by new language
constructs (e.g., by division). A new form of the \verb|assert| can be added,
such that it propagates intermediate solutions to subsequent commands of the
same sort. Support for global constraints can also be developed, but primarily
only as a ,,syntactic sugar`` --- the user could express global constraints
more easily, but internally they would be expanded as if they were expressed
using loops (i.e., as in the current version of the system). Alternative forms
of support for global constraints would require substantial changes in the
\sat-reduction mechanism.

On the lower algorithmic and implementation level, we are planning to
further improve the current version of transformation to \cnf.

In the current version, ground integers are represented by built-in fixed-precision
integers, which is typically sufficient. However, in order to match symbolic
integers, ground integers should be represented by arbitrary (but fixed)
length integers and we are planning to implement that.

Concerning the underlying \sat solvers, currently only two complete
\sat solvers are used. We are planning to integrate additional solvers, since
some solvers are better suited to some sorts of input instances, as the \sat
competitions show. Within this direction of work, we will also analyze
performance of stochastic solvers within \ursa. In addition, we are
exploring potentials of using non-\cnf{} \sat solvers \cite{ThiffaultBW04,MuhammadS06,JainC09}
within \ursa, which avoids the need for transformation to \cnf \cite{milan_todorovic_msc}.
Choosing among available solvers can be automated by using machine learning techniques
for analysis of the generated \sat instances (or even input specifications)
\cite{Hutter07b,NikolicMJ09}. For solving optimization problems, instead of the
existing naive implementation, we are planning to implement more advanced
approaches and to explore the use of MaxSAT and pseudo-Boolean solvers
\cite{HandbookOfSAT2009}.

On the theoretical side, the full operational semantics outlined in this
paper can be formally defined and it could be proved that solutions
produced by the \ursa system indeed meet the specifications and if
there are no produced solutions, then the specifications is inconsistent.
Along with the formal verification (i.e., verification within a proof assistant)
of the \sat solver \argosat used \cite{dpll-correctness,jar-sat-correctness},
that would provide a formal correctness proof of the \ursa system
(which would make it, probably, the first {\em trusted} constraint solver).

In the presented version of \ursa, reducing to \sat is tightly integrated
(and defined by the semantics of the system) in the program execution phase.
An alternative would be as follows: during the program execution phase,
a first-order formula is generated and only before the solving phase it
is translated to a propositional formula. Moreover, the generated formula
would not need to be translated to a propositional formula, but could be
tested for satisfiability by using \smt (satisfiability modulo theory)
solvers (e.g., for linear arithmetic, equality theory, alldifferent theory
etc.) \cite{BSST09HBSAT}. In particular, symbolic computations employed by
the \ursa system are closely related to the theory of bit-vector arithmetic
and to decision procedures for this theory based on ``bit-blasting''
\cite{BrinkmannD02,BryantKOSSB09}. Since solvers for this theory typically
cover all the operators used in \ursa, the theory of bit-vector arithmetic
can be used as an underlying theory (instead of propositional logic) and any
solver for bit-vectors arithmetic can be used as a solving engine. Generally,
the reduction could
be adaptable to \smt solvers available --- if some solver is available, then
its power can be used, otherwise all generated constraints are exported to
propositional logic. This would make the approach more powerful and such
development is the subject of our current work --- the system {\sc ursa major}
will be able to reduce constraints not only to \sat but to different \smt
theories. That system will be a general constraint solver and also a
high-level front-end to the low level {\sc smt-lib} interchange format,
and, further, to all \smt solvers that supports it.
Reduction to the theory of bit-vector arithmetic is firstly explored
in this context \cite{urbiva} and it shows that reducing to bit-vector
arithmetic does not necessarily lead to more efficient solving process
than reducing to \sat (and the same holds for other \smt theories).

With the increased power of the presented system (by using both \sat
and \smt solvers), we are planning to further consider a wide range
of combinatorial, \np-complete problems, and potential one-way functions
and also to apply the \ursa system to real-world problems (e.g., the
ones that are already being solved by translating them to \sat).
Some applications in synthesis of programs are already the subject
of our current work.

For the sake of easier practical usability of the \ursa system, we are
planning to develop a support for integration of \ursa with popular
imperative languages (C, C++, Java).

% ***************************************************************************
% ***************************************************************************
\section{Conclusions}
\label{sec:conclusions}
% ***************************************************************************
% ***************************************************************************

In this paper we described a novel approach for uniform representation and
solving of a wide class of problems by reducing them to \sat. The approach
relies on:
\begin{iteMize}{$\bullet$}
\item a novel specification language that combines features of imperative
and declarative languages --- a problem is specified by a test, expressed
in an imperative form, that some values indeed make a solution to the problem;
\item symbolic computations over unknowns represented by (finite) vectors of
propositional formulae.
\end{iteMize}

The approach is general and elegant, with precisely defined syntax and
induced (by the concept of symbolic execution) semantics of the specification
language. This enables straightforward implementation of the system and it works
as a ``clear box''.
The proposed language is a novel mixture of imperative and declarative
paradigms, leading to a new programming paradigm. Thanks to the language's
declarative aspects --- the problem is described by what makes a solution
and not by describing how to find it --- using the system does not require
human expertise on the specific problem being solved. On the other hand,
specifications are written in imperative form and this gives the following
advantages compared to other modelling languages (all of them are declarative):

\begin{iteMize}{$\bullet$}
\item problem specifications can involve destructive assignments, which
is not possible in declarative languages and this can be essential for
many sorts of problems (e.g., from software verification);
\item modelling problems that naturally involve loops (and nested loops)
is simple and the translation is straightforward;
\item for users familiar with imperative programming paradigm, it should
be trivial to acquire the specification language \ursa (since there are
no specific commands or flow-controls aimed at constraint solving);
\item the user has a fuller control of internal representation of the
problem, so can influence the efficiency of the solving phase.
\item a specification can be taken, almost as-is, from and to languages
such as C (within C, such code would check if some given concrete
values are indeed a solution of the problem).
\item the system can smoothly extend imperative languages like C/C++ or Java,
(as constraint programming extends logic programming).
\item the system can be verified to be correct in a straightforward manner.
\end{iteMize}

In addition, the system \ursa, in contrast to most of (or all) other modelling
systems, supports bit-wise logical operators and constraints involving
modular arithmetic (for the base of the form $2^n$), which is
essential for many applications, and can also enable efficient problem
representation and problem solving.

The proposed approach can be used for solving all problems with Boolean and
numerical unknowns, over Boolean parameters and numerical parameters with finite
domains that can be stated in the specification language
(e.g., the domain of the system is precisely determined by expressiveness of its
specification language).

The search for required solutions of the given problem is performed by
modern \sat solvers that implement very efficient techniques not directly
applicable to other domains. While \sat is already used for solving a
wide range of various problems, the proposed system makes these reductions
much easier and can replace a range of problem-specific tools for
translating problems to \sat. The tool \ursa can be used not only as a
powerful general problem solver (for finite domains), but also as a tool
for generating \sat benchmarks (both satisfiable and unsatisfiable).\footnote{A
large number of \sat instances obtained from cryptanalysis of \des \cite{milan_sesum_msc}
(by using simpler version of the system proposed in this paper), were
part of the \sat Competition corpus
(\url{http://www.cril.univ-artois.fr/SAT09/solvers/booklet.pdf}).}

Concerning weaknesses, \ursa is not suitable for problems where knowledge of
the domain and problem structure are critical and can be
efficiently tackled only by specialized solvers, and this holds for
reduction to \sat generally. Due to its nature, by interfacing \ursa with
standard specification languages like {\sc xcsp} or {\sc MiniZinc}, most
of the \ursa modelling features and power would be lost (e.g., bit-wise
logical operators and destructive updates), while global constraints
supported by these languages would be translated in an inefficient way.
Therefore, it makes no much sense to enable conversion from these
standard languages to \ursa and this makes \ursa a bit isolated system
in the world of constraint solvers or related systems.

In this paper we do not propose:
\begin{iteMize}{$\bullet$}
\item a new \sat-encoding technique --- rather, the \ursa specification
language can be used for different encoding styles;
\item a technique for transforming a \sat formula to conjunctive normal
form --- this step is not a part of the core of the \ursa language and
is not covered by its semantics, so any approach (meeting the simple
specification) can be used; the current technique seem to work well in
practice, while it can still be a subject of improvements.
\item a \sat solver --- rather, \ursa can use any \sat solver (that can
generate all models for satisfiable input formulae); moreover, having several
\sat solver would be beneficial, since some solvers are better suited to
some sorts of problems.
\end{iteMize}

In this paper we also described the system \ursa that implements the proposed
approach and provided some experimental results and comparison with related
systems. They suggest that, although \ursa is not primarily a \csp solver (but
a general system for reducing problems to \sat), the system is, concerning
efficiency,  competitive to state-of-the-art academic and industrial \csp tools.
\ursa is also competitive to other system that translate problem
specifications to \sat. In contrast to most of other constraint solvers,
the system \ursa is open-source and publicly available on the Internet.

For future work, we are planning to extend the system so it can use not
only complete \sat solvers, but also stochastic \sat solvers, non-\cnf{}
solvers, and \smt solvers. We will also work on formal (machine-checkable
by a proof assistant) verification of the system (i.e., show that the \ursa
system always gives correct results) and on extensions of the system
relevant for practical applications.

\section*{Acknowledgement}
This work was partially supported by the Serbian Ministry of Science
grant 174021 and by SNF grant SCOPES IZ73Z0\_127979/1.

The author is grateful to Dejan Jovanovi\'c whose ideas on overloading C++
operators influenced development of the system presented in this paper;
to Filip Mari\'c for valuable discussions on the presented system and for
using portions of his code for shared expressions; to Milan \v{S}e\v{s}um
for portions of his code for performing operations over vectors of
propositional formulae; to Ralph-Johan Back and to the anonymous reviewers
for very detailed and useful comments on a previous version of this paper.

%\bibliographystyle{theapa}
%\bibliographystyle{plain}
%\bibliography{pj}

\end{document}